\documentclass[conference]{IEEEtran}
\IEEEoverridecommandlockouts
\usepackage{cite}
\usepackage{amsmath,amssymb,amsfonts}
\usepackage{graphicx}
\usepackage{textcomp}
\usepackage{xcolor}
\usepackage{color, colortbl}
\usepackage{graphicx}
\usepackage{amsmath}
\usepackage{amsfonts}
\usepackage{import}
\usepackage{pifont}
\usepackage{multirow}
\usepackage{subfigure}
\usepackage{footmisc}
\usepackage{algorithm}
\usepackage[noend]{algpseudocode}
\usepackage{mathtools}
\usepackage{footmisc}
\usepackage{amssymb}
\usepackage{amsthm}
\usepackage{bm}
\usepackage[flushleft]{threeparttable}

\theoremstyle{definition}

\newtheorem{theorem}{Theorem}[]
\newtheorem{lemma}{Lemma}[]
\def\BibTeX{{\rm B\kern-.05em{\sc i\kern-.025em b}\kern-.08em
    T\kern-.1667em\lower.7ex\hbox{E}\kern-.125emX}}
 
\begin{document}

\title{Communication-Efficient Decentralized Learning with Sparsification and Adaptive Peer Selection}

\author{\IEEEauthorblockN{Zhenheng Tang, Shaohuai Shi, Xiaowen Chu
	\IEEEauthorblockA{Department of Computer Science, Hong Kong Baptist University
		\\\{zhtang, csshshi, chxw\}@comp.hkbu.edu.hk}
}}

\maketitle

\begin{abstract}
Distributed learning techniques such as federated learning have enabled multiple workers to train machine learning models together to reduce the overall training time. However, current distributed training algorithms (centralized or decentralized) suffer from the communication bottleneck on multiple low-bandwidth workers (also on the server under the centralized architecture). Although decentralized algorithms generally have lower communication complexity than the centralized counterpart, they still suffer from the communication bottleneck for workers with low network bandwidth. To deal with the communication problem while being able to preserve the convergence performance, we introduce a novel decentralized training algorithm with the following key features: 1) It does not require a parameter server to maintain the model during training, which avoids the communication pressure on any single peer. 2) Each worker only needs to communicate with a single peer at each communication round with a highly compressed model, which can significantly reduce the communication traffic on the worker. We theoretically prove that our sparsification algorithm still preserves convergence properties. 3) Each worker dynamically selects its peer at different communication rounds to better utilize the bandwidth resources. We conduct experiments with convolutional neural networks on 32 workers to verify the effectiveness of our proposed algorithm compared to seven existing methods. Experimental results show that our algorithm significantly reduces the communication traffic and generally select relatively high bandwidth peers.
\end{abstract}

\section{Introduction}
The increasing amount of data plays an important role in the success of modern machine learning applications, and the increasing size of machine learning models, especially deep neural network models, improves the generalization ability. However, larger size of training data and models requires more computing resources to train the model. Distributed learning techniques, such as parallel stochastic gradient descent (PSGD) and its variants, have been widely deployed to train large models by exploiting multiple computing nodes \cite{dean2012large}. The update rule of PSGD with $n$ workers at iteration $t$ is
\begin{equation}
    \mathbf{x}_{t+1}=\mathbf{x}_{t}-\gamma_t\frac{1}{n}\sum_{i=1}^{n}G^i(\mathbf{x}_t),
\end{equation}
where $\mathbf{x}_t$ is the model parameter, $\gamma_t$ is the learning rate, and $G^i(\mathbf{x}_t)$ is the stochastic gradients of worker $i$. Yet, the communication (exchanging gradients or models) between the workers may become the system bottleneck that limits the scalability of the distributed system. There are two types of architectures, centralized and decentralized, to support scalable distributed learning. 
The parameter server (PS) architecture \cite{li2014scaling,li2014communication} is widely applied \cite{zhou2017kunpeng,konevcny2016federated,mcmahan2017communication,hsieh2017gaia,shi2018performance} and is also integrated in popular machine learning frameworks (e.g., TensorFlow \cite{abadi2016tensorflow}). In each iteration, a worker pulls the latest model from the PS and trains the model with its local data. In the original PSGD with PS (PS-PSGD), each worker pushes its gradients to the PS, and the PS updates the model with the average gradients. In the recent federated learning algorithm, FedAvg \cite{konevcny2016federated,mcmahan2017communication}, the workers send their local models to the PS for averaging after several rounds of updates. In both PS-PSGD and FedAvg, the PS and workers suffer from three aspects of communication overheads. First, the PS should send (and receive) models (the model size $N$ could be from millions to billions) to (and from) a certain number of workers (say $n$), which requires $2\times N \times n$ of communication traffic. Second, every worker needs to pull the latest model from the PS, which requires down-stream communication traffic of $N$ in each round. Third, every worker needs to push the local gradients/model to the PS, which requires up-stream communication traffic of $N$. On the PS side, although the communication pressure can be alleviated by deploying multiple PSes \cite{corrado2014training}, there exist many system parameters to tune to achieve good scaling efficiency \cite{yan2015performance,shi2018adag}. On the workers' side, the down- and up-stream communications are also significant for large models. Jakub et al. \cite{konevcny2016federated} propose to use structured or random updates based on the FedAvg (S-FedAvg) algorithm to sparsify the model to reduce the communications on the server and workers. However, S-FedAvg only alleviates the up-stream traffic of the worker, while the communication on the server and down-stream communication on the worker remain to be significant.

The decentralized architecture is an alternative solution for distributed learning. The classical decentralized learning is PSGD with MPI collectives (e.g., all-reduce) \cite{awan2017s,goyal2017accurate,jia2018highly} as MPI has a long history in the HPC community for providing efficient communication primitives \cite{pjevsivac2007performance}. Recently, new communication libraries like NCCL\footnote{\url{https://developer.nvidia.com/nccl}} and Gloo\footnote{\url{https://github.com/facebookincubator/gloo}} have been developed to support high throughput and low latency communication for dense GPU clusters. The all-reduce based methods eliminate the bottleneck of the central server in the PS architecture, but the communication complexity (the bandwidth term) on the worker side is $O(N)$, which could also limit the system scalability.

On one hand, to reduce the communication size of gradients, gradient compression techniques including quantization \cite{alistarh2017qsgd,wen2017terngrad,chen2017adacomp} and sparsification \cite{lin2017deep,shi2019distributed,shi2020communication} can be used in PSGD. The gradient sparsification method is more aggressive than the gradient quantization method in reducing the communication size. For example, Top-$k$ sparsification (TopK-PSGD) \cite{lin2017deep,shi2019aconvergence,karimireddy2019error} with error compensation can zero-out $99\%-99.9\%$ local gradients with little impact on the model convergence while quantization by reducing 32-bit to 1-bit only achieves a maximum of 32$\times$ compression. Although TopK-PSGD can locally zero-out $99\%-99.9\%$ gradients, each worker needs to gather all other $n-1$ workers' sparsified gradients so that the communication complexity of TopK-PSGD is $O(nN/c)$, which is linear to the number of workers. 

\begin{figure}[!ht]
	\centering
	\includegraphics[width=0.98\linewidth]{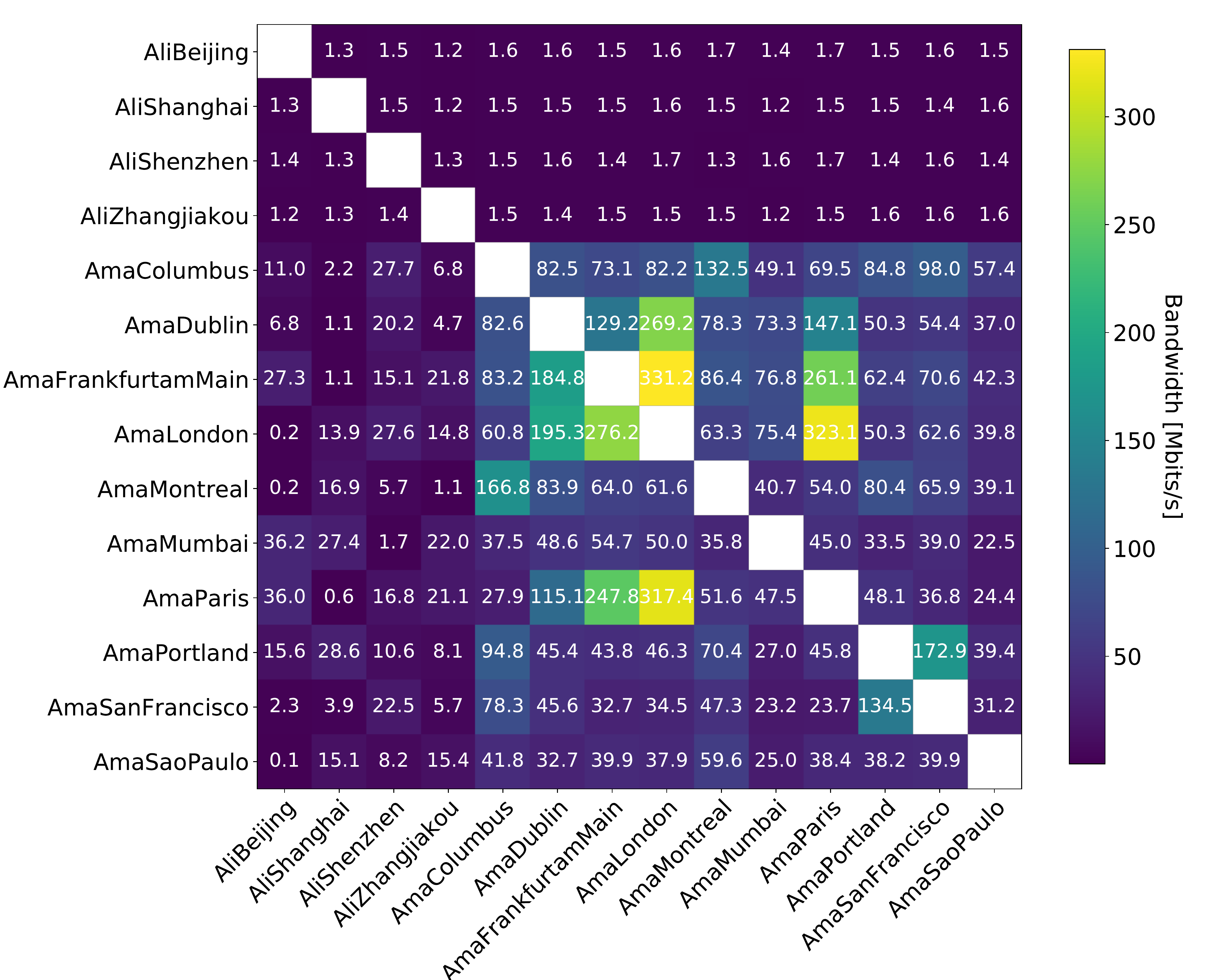}
	\vspace{-10pt}
	\caption{Network speeds between virtual machines located at different cities.}
	\label{fig:networkspeed}
\end{figure}

On the other hand, to reduce the total amount of traffic, Lian et al. \cite{lian2017can} propose the D-PSGD learning algorithm, in which each worker only exchanges the model with some peers instead of all $n-1$ workers. D-PSGD requires a communication complexity of $O(\text{the degree of the network})$. To further reduce the communication traffic on each worker, Tan et al. \cite{DCD-PSGD} propose DCD-PSGD to compress the model to be exchanged between workers. DCD-PSGD to some extent alleviates the communication traffic of the workers, but it has two main limitations: 1) The worker is required to have large memory to store all other workers' models. 2) It requires that the network topology should keep unchanged to guarantee the training convergence. However, on the federated learning setting, the workers are resource-limited and very dynamic, and they may join/leave the training randomly due to the battery power, network connection, network latency, resource availability, etc. As multiple workers may be located on diverse geographical locations, the bandwidth between two workers may also vary. According to our experimental study on network speed tests (as shown in Fig. \ref{fig:networkspeed}) on different cloud service providers located at different cities, the network speed varies from different locations across multiple cities. To the best of our knowledge, there is no distributed training algorithm yet that exploits this bandwidth diversity.

In summary, existing distributed learning algorithms suffer from the communication bottleneck on either the server or the workers. The diversity of bandwidth among workers is under-explored in current decentralized distributed learning algorithms. In this paper, we propose SAPS-PSGD, a communication-efficient distributed learning algorithm with sparsification to reduce the communication traffic on workers and with adaptive peer selection to fully utilize the bandwidth resources between different workers. We theoretically prove that SAPS-PSGD has theoretical convergence guarantees on non-convex smooth problems. The SAPS-PSGD algorithm has the following features: 1) It follows a decentralized architecture without using a PS; instead, it uses a lightweight coordinator for management purpose, which is similar to the BitTorrent tracker. 2) At every communication round, each worker only needs to exchange a highly compressed model with a single peer, which significantly reduces both the up-link and down-link communication traffic. 3) The peers of each worker are dynamically chosen according to their connection bandwidths, which can achieve high usage of the global bandwidth resources. Experimental results show that SAPS-PSGD not only has much lower communication traffic than existing algorithms, but it also achieves better utilization of the bandwidth resources. Our contributions are summarized as follows:
\begin{itemize}
    \item We propose a communication-efficient decentralized distributed learning algorithm named SAPS-PSGD that considers communication efficiency and bandwidth resource utilization.
    \item We theoretically prove that SAPS-PSGD provides convergence guarantees for training machine learning models with non-convex objective functions, and has a consistent convergence rate with PSGD.
    \item We conduct experiments on various  models to verify the convergence performance and the reduction of communication traffic during the training.
\end{itemize}

The rest of the paper is organized as follows. We illustrate the details of our proposed decentralized learning algorithm in Section \ref{sec:algorithm}, followed by the theoretical convergence analysis of the algorithm in Section \ref{sec:analysis}. We demonstrate the experimental study to evaluate the convergence performance and communication efficiency in Section \ref{sec:evaluation}. Related work is introduced in Section \ref{sec:relatedwork}. Finally, we conclude the paper in Section \ref{sec:conclusion}.

\section{Algorithm} \label{sec:algorithm}
In this section, we present our communication-efficient decentralized learning algorithm (SAPS-PSGD). 

\begin{figure}[!ht]
	\centering
	\includegraphics[width=1.0\linewidth]{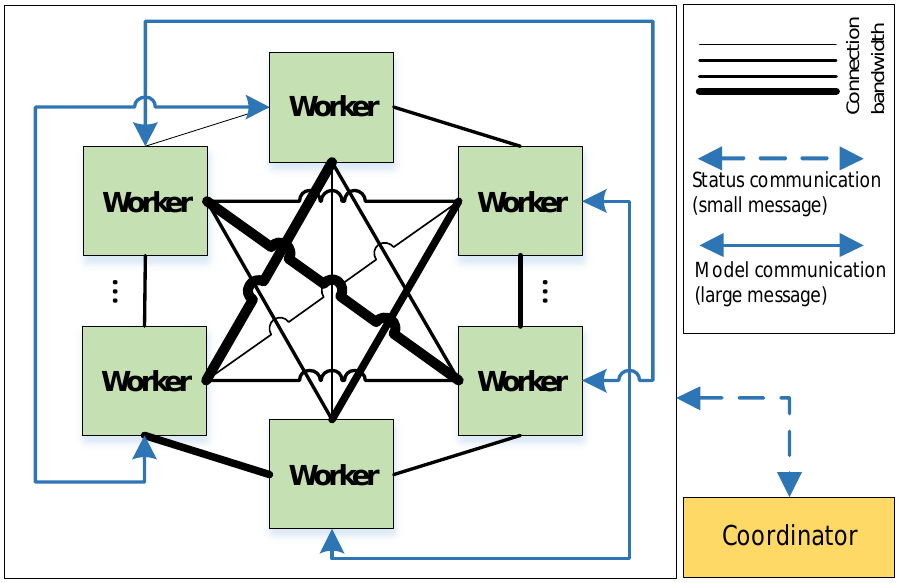}
	\caption{Topology of our decentralized learning algorithm: SAPS-PSGD. The green boxes are training workers who hold local models during training. The yellow box is the coordinator who maintains key information of all workers, e.g., the communication bandwidth of workers. The virtual blue line with arrows indicates the small messages (e.g., training loss) exchange between the coordinator and the workers. The black lines with different width (thicker lines have higher bandwidth) indicate the connection bandwidth between workers. The solid lines with arrows indicate the sparse model exchange between workers. The coordinator prefers to notify the workers to select peers with higher bandwidth to exchange models.}
	\label{fig:topology}
\end{figure}

\subsection{Architecture Overview}
There are two components: Coordinator and Worker on SAPS-PSGD. The architecture overview is shown in Fig. \ref{fig:topology}. 

\textbf{Coordinator.} The coordinator is a central server that manages global information about the training process. Note that the coordinator defined in our framework is not a parameter server that needs to collect model parameters or gradients. The global information contains: model architecture name (e.g., ResNet-50 \cite{he2016deep}), global step $t$, bandwidth matrix $B$ of connected workers, the model exchange matrix $W_t$ and a random seed $s$. At the beginning of training, the coordinator initializes a model training task and distributes the task to all the connected workers. At iteration $t$, the coordinator sends the global information to all participating workers. The pseudo-code of the algorithm on the coordinator side is shown in Algorithm \ref{algo:coordinator}. First, the algorithm will set two nodes $i,j$ connected if the bandwidth $B_{ij}$ between them exceeds a user-defined threshold $B_{thres}$, generating filtered bandwidth matrix $B^*$. Then at iteration $t$, the algorithm generates (Line 4) the gossip matrix $W_t$ satisfying the Assumption \ref{eigenvalueassumption} (the details of the gossip matrix will be shown in Section \ref{subsec:gossipmatrix}) so that each worker can find its peer in the matrix. The coordinator also generates a random seed $s$ (Line 5) for workers to generate the random mask vector $\mathbf{m}_t$ (the details of $\mathbf{m}_t$ will be introduced in Section \ref{subsec:masks}). Then it sends $W_t$, $t$ and $s$ (Line 6) to all the participating workers. After that it waits for the finished messages (say ``ROUND\_END'') of the current round from workers (Line 7). After receiving the notification messages ``ROUND\_END'' from workers, the coordinator continues for the next iteration. Finally, the coordinator receives a final full model from any worker.

\begin{algorithm}[h]
	\caption{SAPS-PSGD at the coordinator}\label{algo:coordinator}
	\small
	\textbf{Input: } $netName, B, T, B_{thres}$
	\begin{algorithmic}[1]
		\small
		\State Initialize connections with workers;
		\State \Call{GetNewConnectedGraph}{$B, B_{thres}$}
		\For{$t=1\rightarrow T$}
		    \State $W_t=$ \Call{GenerateGossipMatrix}{$B, t$}; //Refer to Sec. \ref{subsec:gossipmatrix}
		    \State $s=$ a random number as the seed; 
		    \State \Call{NotifyWorkerToTrain}{$W_t, t, s$};
		    \State \Call{WaitWorkersForCurrentRound}{};
		\EndFor
		\State \Call{CollectFullModelFromOneWorker}{};
		\Procedure{GetNewConnectedGraph}{$B, B_{thres}$}
		    \State $\forall (i,j)\in B$ if $ B_{ij} \ge B_{thres}$: $B^*_{ij}=1 $;
		    \State $\forall (i,j)\in B$ if $ B_{ij} < B_{thres}$: $B^*_{ij}=0 $;
		    \State Return $B^*$;
		\EndProcedure
	\end{algorithmic}
\end{algorithm}

\textbf{Worker.} A worker in the SAPS-PSGD algorithm is defined as the training worker collaborated with other workers. The worker trains a single model with local data using mini-batch SGD, and iteratively communicates with its peer to exchange the sparsified model. The pseudo-code of the algorithm on the worker side is shown in Algorithm \ref{algo:worker}. At the beginning of training, the worker first initializes the connection (Line 1) with the coordinator and initializes (Line 2) the training model with the network architecture ($netName$). From iteration $1$ to $T$, each worker receives message ($W_t$, $t$, $s$) from the coordinator (Line 4), and then runs $SGD$ with local training data (Line 5). Next, each worker generates the same mask vector $\mathbf{m}_t\in \mathbb{R}^{N\times 1}$ with the seed ($s$) to indicate that which components of the model (i.e., sparsification) should be exchanged (Line 6-7) with its peer. The peer is specified from $W_t$ (Line 8). At Line 9, the worker sends its own sparsified parameters ($\widetilde{\mathbf{x}}$) to its peer and receives the sparsified parameters ($\widetilde{\mathbf{x}}_{peer}$) from the peer, and then averages the received parameters with the local one at Line 10. At the end of the iteration (Line 11), the worker sends a ``ROUND\_END'' message to the coordinator to notify that it has finished current round.

\begin{algorithm}[h]
	\caption{SAPS-PSGD at worker $p$}\label{algo:worker}
	\small
	\textbf{Input: } $netName, T, rank, D_p, L$
	\begin{algorithmic}[1]
		\small
		\State Initialize a connection with the coordinator;
		\State $net=$ Initialize the model with $netName$;
		\For{$t=1\rightarrow T$}
		    \State $W_t, t, s=$ \Call{RecieveMsgFromCoordinator}{};
		    \State \Call{SGD}{$net, D_p, L$}
		    \State $\mathbf{m}_t=$ \Call{GenerateRandomMask}{$s$};
		    \State $\widetilde{x}=net.x\circ \mathbf{m}_t$;
		    \State $peer=$ $W_t[rank]$;
		    \State $\widetilde{x}_{peer}=$ \Call{ExchangeModelWithPeer}{$\widetilde{x}, peer$};
		    \State $net.x=net.x\circ \neg \mathbf{m}_t+\widetilde{x}_{peer}$;
		    \State \Call{SendEndOfRoundToCoordinator}{};
		\EndFor
		\State \Call{CollectModelFromOneWorker}{};
		\Procedure{SGD}{$net, D_p, L$}
		    \State $[d, y]=$ Sample a mini-batch of data from $D_p$;
		    \State $loss=$ \Call{ComputeLoss}{$net.x, d, y$};
		    \State $net.x=net.x- \gamma \nabla net.x$
		\EndProcedure
	\end{algorithmic}
\end{algorithm}

\subsection{Model Sparsification} \label{subsec:masks}
We denote the model of one worker at iteration $t$ by $\mathbf{x}_t\in \mathbb{R}^{N\times 1}$. At the $t^{th}$ communication round, the model is sparsified as $\widetilde{\mathbf{x}}_t$ by zeroing-out most of its components (e.g., $99\%$) in $\mathbf{x}_t$. Then the worker sends $\widetilde{\mathbf{x}}_t$ to its peer. Thus, we can generate a mask vector $\mathbf{m}_t\in \mathbb{R}^{N\times 1}$ whose elements are either $1$ or $0$ to achieve $\widetilde{\mathbf{x}}_t$, i.e, 
\begin{equation}
    \widetilde{\mathbf{x}}_t = \mathbf{x}_t \circ \mathbf{m}_t,
\end{equation}
where $\circ$ is the Hadamard product operator on vectors or matrices. As a result, the zeroed elements do not need to be communicated across the network, and thus reduce the communication cost.

The mask vector $\mathbf{m}_t$ is randomly generated at the worker side with a random $seed$ received from the server at iteration $t$ such that all workers generate the same mask matrix at the current state. Therefore, at each communication round, the exchanged components (indices on the model vector) of the model are the same between any two workers. Note that we do not exploit the top-k components as Top-k selection is not efficient \cite{shi2019understanding} and it is hard to guarantee the convergence. To generate the mask matrix with a specific compression ratio $c$, we need to randomly generate the matrix such that its $N(1-1/c)$ elements are zeros and the other $N/c$ elements are ones. We use the Bernoulli distribution with a probability of $1/c$ to generate the mask matrix, i.e.,
\begin{equation}\label{equ:pq}
    \mathbf{m}_t^{[j]}=
    \begin{cases}
1, & \text{with probability} \ p=1/c \\
0, & \text{with probability} \ q =1-1/c
\end{cases}.
\end{equation}
We use $M_t \in \mathbb{R}^{N \times n}$ to denote masks of all workers at iteration $t$, whose columns are the same vector $\mathbf{m}_t$.

As $\widetilde{\mathbf{x}}_t$ contains at least $N(1-1/c)$ zero elements, the worker sends the non-zero elements (less than $N/c$) to its peer to save the communication traffic. Therefore, the communication traffic on one worker to receive and send the sparsified models is less than $2N/c$ at each communication round, which is much smaller than the D-PSGD \cite{lian2017can} and DCD-PSGD \cite{DCD-PSGD} algorithms.


\subsection{Gossip Matrix} \label{subsec:gossipmatrix}
The gossip algorithms describe a class of distributed average problems \cite{gossipalgorithms,RandomizedGossipAlgorithms}. Every worker $i$ possesses its data and exchange it with other workers based on a gossip matrix, which can be formulated as following:
\begin{equation}
    {X}_{t} = {X}_{t-1} W_{t-1},
\end{equation}
where the ${X}_t$ represents the data owned by workers, i.e., the $i$-th column is the data owned by worker $i$.

The spectral gap $\rho_{sp} < 1$ of the gossip matrix $W_t$ is required to guarantee that the algorithm can reach a consensus and convergence in decentralized distributed learning \cite{lian2017can}\cite{DCD-PSGD}, in which the consensus means that the running average of $\mathbb{E} \| \frac{\sum_{i=1}^n x_i}{n} - x_i\|^2$ converges to $0$.

The gossip matrix and its generation in our SAPS-PSGD algorithm is different from the previous works \cite{lian2017can,DCD-PSGD} in two aspects. 

First, it is known that the faster the algorithm reaches the consensus, the smaller spectral gap \cite{RandomizedGossipAlgorithms}. One can add more connections in the graph to achieve faster consensus, but it would introduce more communications. So there exists a trade-off between communication efficiency and the time to achieve consensus. In \cite{lian2017can}\cite{DCD-PSGD}, the connected peers of a worker are selected to be the communicated peers and the communication topology at every iteration is required to be a \textbf{Connected Graph} to satisfy $\rho_{sp} < 1$ so that each worker should communicate with at least two other peers. In SAPS-PSGD, we use a single-peer communication scheme (each worker only communicates with a single peer), so each row in our gossip matrix has only two non-zero elements. Consequently, the communication traffic at each worker is at least two times smaller than D-PSGD \cite{lian2017can} and DCD-PSGD \cite{DCD-PSGD}. Note that $\rho_{sp}$ of $W_t$ in SAPS-PSGD is not required be smaller than $1$, but we should guarantee that the second largest eigenvalue $\rho$ of $\mathbb{E}(W_t^TW_t) $ is smaller than $1$. To guarantee this property, all possible communication edges (\textbf{PC edges}) that have a possibility to be chosen should construct a connected graph \cite{RandomizedGossipAlgorithms}.

Second, under the configuration that each worker communicates with no more than two peers in \cite{lian2017can}\cite{DCD-PSGD}, the best topology that can most efficiently spread information is the ring topology. However, choosing the best ring-topology with diverse link bandwidths is to find a Hamilton Cycle which is a classical NP-Complete problem \cite{karp1972reducibility}. One may consider to choose one best graph traversal to exploit the highest bandwidth, which is easier to obtain but it could loses the connection between the start and the end of the traversal and decreases the speed of information propagation. Our gossip matrix generation algorithm achieves a balance that can generate a good communication topology without losing much efficiency of information propagation.

To summarize, our gossip matrix only requires each worker to communicate with one peer, and at the same time considering better bandwidth exploitation. Therefore, we generate it following the new assumption of the second eigenvalue similar to \cite{RandomizedGossipAlgorithms,gossipalgorithms,randomrumorspreading} but different from that in \cite{DCD-PSGD}. There is a key property of our random gossip matrices $W_t$ (i.i.d. and $\rho<1$). That is, for any row vector sequence $\mathbf{x_{t}} \in \mathbb{R}^{1\times n}$ defined as
\begin{equation*}
    \mathbf{x}_{t} = \mathbf{x}_{t-1} W_{t-1}, s<t,
\end{equation*}
we have 
\begin{equation}\label{gossipconsensus}
    \mathbb{E}_{W_s, W_{s+1}, \ldots W_{t-1}}{\left \| \mathbf{x}_{t} - \overline{\mathbf{x}}_{t}\mathbf{1}_n^{\top}   \right \|}^{2} \leq \rho^{2\left( t-s \right)} {\left \| \mathbf{x}_{s} - \overline{\mathbf{x}}_{s}\mathbf{1}_n^{\top}   \right \|}^{2},
\end{equation}
where $\rho$ is the second largest eigenvalue of $E[{W_t}^T W_t]$ \cite{RandomizedGossipAlgorithms}.

It is known that the connection speeds in geo-distributed workers are different as shown in Fig. \ref{fig:networkspeed}. If the worker randomly selects the peer, then the model transmission between some workers could be very slow. To address the problem, we propose a novel gossip matrix generation method according to the communication speed of each pair of workers, which tries to maximize the network resource utilization and thus more efficient communications, at the same time ensuring all PC edges can construct a connected graph. Assume that the coordinator has the communication speed information\footnote{In practice, the communication speed information is measured by each pair of peers and regularly reported to the coordinator. } between any two workers with a matrix $B$, where $B_{ij}$ is the communication speed between worker $i$ and $j$. And we set $B_{ij}=B_{ji}=min(B_{ij}, B_{ji})$ as the communication bottleneck is decided by the slow one.

Our gossip matrix generation algorithm is shown in Algorithm \ref{algo:gossipmatrix}. To satisfy Assumption \ref{eigenvalueassumption}, we define a communication iteration gap $T_{thres}$, and a timestamp matrix $R$ to record the communication at every iteration. We call the edge $(i, j)$ satisfying $ R_{ij} > t - T_{thres}$ as the ``recently connected'' (RC) edge. At first, the algorithm will judge if all RC edges can construct a connected graph (line 1). If they can, the algorithm finds the maximum match using the filtered bandwidth matrix $B^*$ (line 2 and 5). Otherwise, the algorithm finds all connected sub-graphs constructed from RC edges, and then chooses edges in $B$ that connect all sub-graphs to generate a new matrix $E$ which is used to find the maximum match (line 4 and line 5). After doing the first match, there is possibility that some workers haven't been matched, if so, the algorithm will do maximum match with unmatched workers again without considering bandwidth (line 6, 7 and 8). Then, all workers will be matched (line 9). This perfect match indicates which peer to exchange the sparsified model for a worker, then the coordinator generates (line 10) the gossip matrix $W_t\in \mathbb{R}^{n\times n}$. It is obvious that $W_t$ is a doubly stochastic matrix. Here, we exploit the blossom algorithm \cite{edmonds_1965} to solve the problem of maximum match in a general graph. And by randomly starting from different node in a graph, we implement the RamdomlyMaxMatch function.

\begin{algorithm}[h]
	\caption{GenerateGossipMatrix}\label{algo:gossipmatrix}
	\small
	\textbf{Input: } $B, B^*, R, T_{thres}, n, t$ \\
	\textbf{Output: } $W_t$ 
	\begin{algorithmic}[1]
        \If{\Call{IfConnected}{$R, T_{thres}, t$}}
            \State $E = B^*$;
        \Else
            \State $E$ = \Call{GetOverTimeMatrix}{$R, T_{thres}, t$};
        \EndIf
		\State $match$=\Call{RandomlyMaxMatch}{$E$};
        \If{\Call{Len}{$match$} $\ne$ $n/2$}
    	    \State $E$ = \Call{GetUnmatch}{$B, match$};
    	    \State $match^\prime$=\Call{RandomlyMaxMatch}{$E$};
    	\EndIf
        \State $match$ = $match$ + $match^\prime$;
		\State $W_t$=\Call{GenerateW}{$match$};
        \State Return $W_t$;
        \Procedure{IfConnected}{$R, T_{thres}, t$}
            \State $\forall (i, j)$ if  $R_{ij} > t - T_{thres}$ : \ $Q_{ij}=Q_{ji}=1$;
            \State Return \Call{IfStronlyConnected}{$Q$};
		\EndProcedure
        \Procedure{GetOverTimeMatrix}{$R, T_{thres}, t$}
            \State $\forall (i, j)$ if $R_{ij} > t - T_{thres}$ : \ $Q_{ij}=Q_{ji}=1$;
            \State $S$=\Call{FindConnectedSubgraph}{$Q$};
            \State $E_{ij} = 1, \forall i\in S_k, j\in S_l, k\ne l$;
            \State Return $E$;
		\EndProcedure
		\Procedure{GetUnmatch}{$B, match$}
		    \State $\forall (i, j)\in B$  if $i\not\in match $and$ j\not\in match  : \ E_{ij}=E_{ji}=1$;
		    \State Return $E$;
		\EndProcedure
		\Procedure{GenerateW}{$B, match$}
            \State $\forall (i, j) \in match \ W_{t\_ij}=W_{t\_ji}=1/2$;
            \State $\forall (i, j) \not\in match$ and $i\ne j \ W_{t\_ij}=0 $;
            \State $\forall (i, j) if i==j \ W_{t\_ij}=1/2$;
		\EndProcedure
	\end{algorithmic}
\end{algorithm}

\subsection{Communication Complexity Analysis}
Assume that there are $n$ workers and a coordinator participating in training a model whose size is $N$ with $T$ iterations to achieve a converged model, the communication cost of SAPS-PSGD of coordinator is $N$ as it only receives the final model from a worker. At each round, each worker sends and receives the sparsified model with size of $N/c$, so the communication cost of the workers is $2N\times T/c$. Comparison with other traditional methods is shown in Table \ref{table:comm}. It can be seen that our algorithm not only has the lowest communication cost at the worker side, but also considers the bandwidth of workers to support more efficient communications.

\begin{table}[!ht]
	\centering
	\caption{Communication cost comparison of different algorithms.}
	\label{table:comm}
	\begin{threeparttable}
	\addtolength{\tabcolsep}{-1.5pt}
	\begin{tabular}{|l|c|c|c|c|c|}
		\hline
		Algorithm & Server Cost & Worker Cost & SP. & C.B. & R. \\\hline\hline
		PS-PSGD & $2N n T$ & $2N T$ & \ding{55} & \ding{55}&\ding{55}  \\\hline
		PSGD (all-reduce) & - & $2N T$ & \ding{55} & \ding{55}&\ding{55}  \\\hline
		TopK-PSGD \cite{renggli2019sparcml} & - & $2n(N/c)T$ & \ding{51} & \ding{55} &\ding{55}  \\\hline
        FedAvg \cite{mcmahan2016communication} & $2N n T$ & $2N T$ & \ding{55} & \ding{55} &\ding{55}  \\\hline
		S-FedAvg \cite{konevcny2016federated} & $(N+2N/c)  n  T$ & $(N+2N/c)  T$ & \ding{51} & \ding{55} &\ding{55} \\\hline
		D-PSGD \cite{lian2017can} & $N$ & $4n_{p}N T$& \ding{55} & \ding{55} &\ding{55} \\\hline
		DCD-PSGD \cite{DCD-PSGD} & $N$ & $4n_{p}(N/c) T$&\ding{51} & \ding{55}&\ding{55} \\\hline
		SAPS-PSGD & $N$ & $2(N/c) T$ & \ding{51} & \ding{51}&\ding{51}\\\hline
	\end{tabular}
    \begin{tablenotes}
    	\item Note: ``SP.'' indicates if supporting model/gradient sparsification. ``C.B.'' indicates if considering the bandwidth of clients, and ``R.'' indicates if robustly adapting to the network dynamics. $c$ is the compression ratio and $n_{p}$ is the maximum number of neighbors of one worker and $n_{p}>1$.
    \end{tablenotes}
	\end{threeparttable}
\end{table}

\section{Convergence Analysis} \label{sec:analysis}
For ease of presentation, we summarize the frequently used notations as follows:
\begin{itemize}
    \item $\mathbf{1}_n$=$\left[1 \ 1 \ldots 1 \ 1 \right]^\top \in \mathbb{R}^n$: A full-one column vector.
    \item $\mathbf{e}^{(i)}_n$=$\left[0 \ 0 \ldots 1 \ldots 0 \ 0 \right]^\top \in \mathbb{R}^n$: A column vector whose $i$-th element equals to $1$.
    \item $\mathbf{e}^{\left[ i\right]}_n$=$\left[0 \ 0 \ldots 1 \ldots 0 \ 0 \right] \in \mathbb{R}^n $ : A row vector whose $i$-th element equals to $1$.
    \item $\mathbf{x}^{(i)}_t$=$X_t \mathbf{e}^{(i)}_n \in \mathbb{R}^{N\times 1} $: A column vector of matrix $X_t$.
    \item $\mathbf{x}^{\left[ j\right]}_t$=$\mathbf{e}^{\left[ i\right]}_N  X_t \in \mathbb{R}^{1\times n}$ : A row vector of matrix $X_t$.
    \item $\| \cdot \|$: $l_2$ norm.
    \item $ \bigsqcup\limits_{s=1}^{t} $ : Hadamard product from $1$ to $t$.
\end{itemize}

Formally, the iterative learning of SAPS-PSGD is a decentralized optimization problem of the following objective:
\begin{equation}
      \min_{\mathbf{x} \in \mathbb{R}^N} f(\mathbf{x}) =  \frac{1}{n} \sum_{i=1}^{n} \underbrace{\mathbb{E}_{\xi \sim \mathcal{D}_i} F_i(\mathbf{x};\xi)}_{=:f_i(\mathbf{x})}, 
\end{equation}
where $n$ is the number of workers, $\mathcal{D}_i $ and $F_i$ are the data distribution and loss function of worker $i$, respectively. We use the similar definitions with \cite{DCD-PSGD}:
\begin{align*}
     X&:=[\mathbf{x}^{(1)},\mathbf{x}^{(2)},\ldots,\mathbf{x}^{(n)}] \in \mathbb{R}^{N\times n}, \\
     G(X;\xi)& :=[\nabla F_1(\mathbf{x}^{(1)};\xi^{(1)}),\ldots,\nabla F_n(\mathbf{x}^{(n)};\xi^{(n)})], \\
     \nabla f(\overline{X})&:=\sum_{i=1}^{n}\frac{1}{n}\nabla f_i(\frac{1}{n}\sum_{i=1}^{n}\mathbf{x}^{(i)}) ,\\
    \overline{\nabla f}(X)& :=\mathbb{E}_\xi G(X;\xi) \frac{\mathbf{1}}{n} = \frac{1}{n}\sum_{i=1}^n \nabla f_i(\mathbf{x}^{(i)}).
\end{align*}
Then our SAPS-PSGD algorithm can be generalized into the below form:
\begin{equation}\label{equ:problem}
    {X}_{t+1} = {X}_{t} \circ \lnot {M}_{t} + {X}_{t} \circ {M}_{t} \times {W}_{t} - \gamma_{t}G({X}_{t};\xi_{t}).
\end{equation} 
Our goal is to prove that ${X}_{t}$ converges and the convergence rate is the same as PSGD \cite{dekel2012optimal}. Note that the combination of Hadamard product and matrix product dose not obey the law of combination, which means that one should do products according to its order. Throughout this paper, we omit ``$\times$'' in matrix product expressions. We can re-write Eq. (\ref{equ:problem}) to
\begin{align} \label{iterationXt}
    {X}_{t} & = {X}_0 \bigsqcup\limits_{s=0}^{t} ( \lnot {M}_{s} + {M}_{s} {W}_{s} ) \notag \\
    & - \sum_{s=0}^{t-1} \gamma_{s}G({X}_{s};\xi_{s}) \bigsqcup\limits_{r=s+1}^{t-1} ( \lnot {M}_{r} + {M}_{r} {W}_{r} ).
\end{align}

\subsection{Assumptions} \label{subsec:assumptions}
We make the following general assumptions for the optimization problem.
\begin{enumerate}
    \item $\forall i, f_i(\cdot)$ is with L-Lipschitzian gradients.
    \item $\forall t, W_t$ is doubly stochastic and i.i.d.
    \item \label{eigenvalueassumption} $\forall t, \mathbb{E}\left({W_t}^T W_t\right)$ has the second largest eigenvalue $\rho < 1$.
    \item The variance of stochastic gradients is bounded, i.e.,
    \begin{align*}
        & \mathbb{E}_{\xi \sim \mathcal{D}_i} \| \nabla F_i(\mathbf{x};\xi)-\nabla f_i(\mathbf{x}) \|^2 \leq \sigma^2 , \forall i, \forall \mathbf{x} \\
        & \frac{1}{n} \sum_{i=1}^n \|\nabla f_i(\mathbf{x})- \nabla f(\mathbf{x}) \|^2 \leq \zeta^2 , \forall i, \forall \mathbf{x}
    \end{align*}
\end{enumerate}
\subsection{Useful Facts}
In this subsection, we derive some useful facts that will be used in the following proofs.
\begin{lemma}\label{lemma:maskorderchange}
For any matrix ${X}_t \in \mathbb{R}^{N \times n}$ calculated by the following formulation
\begin{equation}\label{multimatrixproduct}
     {X}_t = {X}_0  \underbrace{\circ {M}_0  {W}_0  \circ {M}_1  \circ {M}_2  {W}_1 \circ {M}_3   {W}_2 \circ {M}_4 \ldots }_ {\text{$t$ mask matrices and $k$ gossip matrices}},
\end{equation}
for any multiplication order, we can rewrite 
\begin{equation}\label{ordermatrixproduct}
    {X}_t = {X}_0 \bigsqcup\limits_{i=0}^{t-1} {M}_i \prod_{i=0}^{k-1} {W}_i.
\end{equation}
\begin{proof}
For any matrix $A \in \mathbb{R}^{N \times n} $, we define two matrices $ Y=A \circ M  W$ and $ Z = A  W \circ M $. Here $M^{(i,1)}=M^{(i,2)}= \ldots =M^{(i,n)}$, and the $i$-th raw and $j$-th column element of $Y$ and $Z$ have the following relationship:
\begin{align*}
    Y^{(i,j)} =& \sum_{k=1}^{n}A^{(i,k)}M^{(i,k)} W^{(k,j)} = \sum_{k=1}^{n}A^{(i,k)}M^{(i,j)}W^{(k,j)} \\
    =& \left( \sum_{k=1}^{n}A^{(i,k)}W^{(k,j)}\right) M^{(i,j)} = Z^{(i,j)},
\end{align*}
which indicates $Y=Z$, i.e., $A \circ M  W=A  W \circ M $. It means that we can exchange the product order between $M$ and $W$. Rearranging the order of (\ref{multimatrixproduct}), and putting all matrix products with $W_i$ at the end of the equation, we can obtain (\ref{ordermatrixproduct}).
\end{proof}
\end{lemma}

\begin{lemma}\label{sparsegossip}
Given any row vector sequence $\mathbf{x}_t \in \mathbb{R}^n$ defined as follows,
\begin{equation}
    \mathbf{x}_t = \mathbf{x}_s \bigsqcup\limits_{r=s}^{t-1} ( \lnot \mathbf{m}_{r} + \mathbf{m}_{r}  {W}_{r} ),
\end{equation}
we have 
\begin{align*}
    & \mathbb{E}_{s\ldots (t-1)}{ \| \mathbf{x}_{t} - \overline{\mathbf{x}}_{t}\mathbf{1}_n^{\top} \|}^{2} = \mathbb{E}_{s\ldots (t-1)}{ \| \mathbf{x}_{t} - \mathbf{x}_{t}\frac{\mathbf{1}_n \mathbf{1}_n^{\top}}{n}   \|^{2}} \notag \\
    =&  \mathbb{E}_{s\ldots (t-1)}{ \| \mathbf{x}_{t}(\mathbf{I} - \frac{\mathbf{1}_n \mathbf{1}_n^{\top}}{n} ) \|^{2}} \leq (q + p \rho^2)^{\left( t-s \right)} {\left \| \mathbf{x}_{s} - \overline{\mathbf{x}}_{s}\mathbf{1}_n^{\top}   \right \|}^{2},
\end{align*}
where $ \mathbb{E}_{s\ldots (t-1)}$ represents $\mathbb{E}_{W_s\ldots W_{t-1}, \mathbf{m}_s \ldots\mathbf{m}_{t-1} }$, and $p$ and $q$ are defined in Eq. (\ref{equ:pq}).
\begin{proof} Taking expectation and under the assumption that $W_r$ is i.i.d, we have
\begin{align}  
    & \mathbb{E}_{s\ldots (t-1)} { \| \mathbf{x}_{t} - \overline{\mathbf{x}}_{t}\mathbf{1}_n^{\top}    \|}^{2} =\mathbb{E}_{s\ldots (t-1)}{ \| \mathbf{x}_{t}(\mathbf{I} - \frac{\mathbf{1}_n \mathbf{1}_n^{\top}}{n} ) \|}^{2} \notag \\
    = & \mathbb{E}_{s\ldots (t-1)}  \| \mathbf{x}_s \bigsqcup\limits_{r=s}^{t-1} ( \lnot \mathbf{m}_{r} + \mathbf{m}_{r}  {W}_{r} ) (\mathbf{I} - \frac{\mathbf{1}_n \mathbf{1}_n^{\top}}{n} )   \|^{2}\notag \\
    =& \sum_{k=s}^{t} \| \mathbf{x}_s W^{k-s}  - \mathbf{x}_s W^{k-s} \frac{\mathbf{1}_n \mathbf{1}_n^{\top}}{n}   \|^{2} C^{k-s}_{n} p^{k-s}q^{t-k} \label{expect_decompose}\notag \\
    =& \sum_{k=s}^{t} \| \mathbf{x}_{s} - \overline{\mathbf{x}}_{s}\mathbf{1}_n^{\top}  \|^{2} \rho^{2(k-s)}C^{k-s}_{n} p^{k-s}q^{t-k} \text{(using Eq. (\ref{gossipconsensus}))} \notag \\
    =& \| \mathbf{x}_{s} - \overline{\mathbf{x}}_{s}\mathbf{1}_n^{\top}  \|^{2} (\sum_{k=0}^{t-s}C^{k}_{n} \rho^{2r} p^{k}q^{t-s-k}) \notag \\
    =& (q + p \rho^2)^{\left( t-s \right)} \|\mathbf{x}_{s} - \overline{\mathbf{x}}_{s}\mathbf{1}_n^{\top}  \|^{2},
\end{align}
which completes the proof.
\end{proof}
\end{lemma}

\subsection{Consensus Analysis}\label{subsec:consensus}
Before bounding the convergence of SAPS-PSGD, we first prove that SAPS-PSGD can reach consensus.


\begin{theorem}\label{theorem:consensus}
    Under the assumptions defined in Section \ref{subsec:assumptions}, if $X_t$ is iteratively updated by Eq. (\ref{iterationXt}), then we have
\begin{multline}
    \sum_{t=1}^{T}\sum_{i=1}^{n}\mathbb{E}\| \mathbf{x}^{(i)}_t  - \overline{{X}_{t}} \|^2  \leq  \frac{2}{1-(q+p\rho^2)} \| X_0 - \overline{{X}_{0}} {\mathbf{1}_n}^\top \|^2_F  \\
    +\frac{2}{(1 - (q+p\rho^2)^{\frac{1}{2}})^2} \sum_{t=1}^{T} \gamma_{t}^2 \mathbb{E}\|G(X_t;\xi_{t}) \|^2_F,
\end{multline}
where $\overline{X} $=$X\frac{\mathbf{1}_n}{n}$.
\begin{proof}
From Eq. (\ref{iterationXt}), we have
\begin{align}\label{xtxbar}
     & \sum_{i=1}^{n}\mathbb{E}\| \mathbf{x}^{(i)}_t  - \overline{{X}_{t}} \|^2 = \sum_{i=1}^{n}\mathbb{E}\| X_t \mathbf{e}^{(i)}_n  - {X}_{t} \frac{\mathbf{1}_n }{n} \|^2 \notag \\
    = & \mathbb{E}\| X_t  - {X}_{t} \frac{\mathbf{1}_n {\mathbf{1}_n}^\top}{n} \|^2_F = \sum_{j=1}^{N}\mathbb{E}\| \mathbf{e}^{\left[ i\right]}_N X_t (\mathbf{I} - \frac{\mathbf{1}_n {\mathbf{1}_n}^\top}{n}) \|^2 \notag\\
    \leq & 2\sum\limits_{j=1}^{N}(\|\mathbf{x}^{\left[ j\right]}_0 - \mathbf{x}^{\left[ j\right]}_0 \frac{\mathbf{1}_n {\mathbf{1}_n}^\top}{n} \|^2 (q + p \rho^2)^t + \mathbb{E} \|\sum_{s=0}^{t-1}\notag \\   
    & \underbrace{\gamma_{s}G^{\left[ j\right]}({X}_{s};\xi_{s}) \bigsqcup\limits_{r=s+1}^{t-1}  ( \lnot \mathbf{m}^{\left[ j\right]}_{r} + \mathbf{m}^{\left[ j\right]}_{r}  {W}_{r} )(\mathbf{I} - \frac{\mathbf{1}_n {\mathbf{1}_n}^\top}{n})}_{Q1}  \|^2),
\end{align}
where the last step is from expanding the $\mathbb{E}\| \mathbf{e}^{\left[ i\right]}_N X_t (\mathbf{I} - \frac{\mathbf{1}_n {\mathbf{1}_n}^\top}{n}) \|^2$ and then using Lemma \ref{sparsegossip}. For convenience, we use $H_s^{\left[ j\right]}$ to denote $Q1$.
Also by using Lemma \ref{sparsegossip}, we have
\begin{align}\label{Hs2}
    & \mathbb{E} \| H_s^{\left[ j\right]} \|^2 \notag \\
    =&  \mathbb{E} \| \gamma_{s}G^{\left[ j\right]}({X}_{s};\xi_{s}) \bigsqcup\limits_{r=s+1}^{t-1}  ( \lnot \mathbf{m}^{\left[ j\right]}_{r} + \mathbf{m}^{\left[ j\right]}_{r} {W}_{r} )(\mathbf{I} - \frac{\mathbf{1}_n {\mathbf{1}_n}^\top}{n}) \|^2 \notag\\
    =&  \mathbb{E}\| \gamma_{s}G^{\left[ j\right]}({X}_{s};\xi_{s}) - \gamma_{s}G^{\left[ j\right]}({X}_{s};\xi_{s}) \frac{\mathbf{1}_n {\mathbf{1}_n}^\top}{n} \|^2 (q + p \rho^2)^{t-s-1} \notag \\
    \leq &   \mathbb{E}\| \gamma_{s}G^{\left[ j\right]}({X}_{s};\xi_{s})\|^2 (q + p \rho)^{t-s-1}.
\end{align}
We further bound $\mathbb{E} \| H_s^{\left[ j\right]}\| \|H_z^{\left[ j\right]}\|$. Taking expectation on time $z+1$ to time $t-1$, we obtain
\begin{align} \label{expectHSHZ}
    & \mathbb{E} \| H_s^{\left[ j\right]}\| \|H_z^{\left[ j\right]}\| \notag \\
    =&   \sum_{k=0}^{t-1-z} \mathbb{E}\| \gamma_{s}G^{\left[ j\right]}({X}_{s};\xi_{s}) (\mathbf{I} - \frac{\mathbf{1}_n {\mathbf{1}_n}^\top}{n}) \|\rho^k (q+p\rho)^{(z-s)}  \notag \\
    &  \cdot \| \gamma_{z}G^{\left[ j\right]}({X}_{z};\xi_{z})(\mathbf{I} - \frac{\mathbf{1}_n {\mathbf{1}_n}^\top}{n}) \|\rho^k C^{k}_{t-1-z}p^k q^{t-1-z-k} \notag \\
    \leq&  \mathbb{E}\| \gamma_{s}G^{\left[ j\right]}({X}_{s};\xi_{s}) \| \cdot \| \gamma_{z}G^{\left[ j\right]}({X}_{z};\xi_{z}) \|  \notag \\
    & \cdot (q+p\rho)^{(z-s)}  (\sum_{k=0}^{t-1-z} \rho^{2k} C^{k}_{t-1-z}p^k q^{t-1-z-k}) \notag \\
    =&  \mathbb{E}\| \gamma_{s}G^{\left[ j\right]}({X}_{s};\xi_{s}) \| \cdot \| \gamma_{z}G^{\left[ j\right]}({X}_{z};\xi_{z}) \| \notag \\
    &  \cdot  (q+p\rho)^{(z-s)}  (q+p\rho^2)^{(t-1-z)}\notag\\
    \leq &  \mathbb{E}\| \gamma_{s}G^{\left[ j\right]}({X}_{s};\xi_{s}) \| \cdot \| \gamma_{z}G^{\left[ j\right]}({X}_{z};\xi_{z}) \| \notag \\
    &  \cdot (q+p\rho)^{\frac{1}{2}(t-1-s)} (q+p\rho)^{\frac{1}{2}(t-1-z)} .
\end{align}

Combining  Eq. (\ref{Hs2}) and (\ref{expectHSHZ}), we have 
\begin{align}\label{HSHZ}
    & \mathbb{E} \|\sum_{s=0}^{t-1} H_s^{\left[ j\right]}  \|^2 \notag = \sum_{s=0}^{t-1} \mathbb{E} \| H_s^{\left[ j\right]} \|^2  + 2\sum_{s<z}^{t-1}\mathbb{E} \langle H_s^{\left[ j\right]}, H_z^{\left[ j\right]}   \rangle\notag \\    
    \leq &   \sum_{s=0}^{t-1} \mathbb{E} \| H_s^{\left[ j\right]} \|^2  + 2\sum_{s<z}^{t-1}\mathbb{E} \| H_s^{\left[ j\right]}\| \|H_z^{\left[ j\right]}\| \notag \\
    \leq &   \sum_{s=0}^{t-1} \mathbb{E}\| \gamma_{s}G^{\left[ j\right]}({X}_{s};\xi_{s})\|^2 (q + p \rho)^{t-s-1}   \notag \\  
    & + 2\sum_{s<z}^{t-1} \mathbb{E}\| \gamma_{s}G^{\left[ j\right]}({X}_{s};\xi_{s}) \|   \cdot \| \gamma_{z}G^{\left[ j\right]}({X}_{z};\xi_{z})\| \notag \\
    &  \cdot (q+p\rho)^{\frac{1}{2}(t-1-s)} (q+p\rho)^{\frac{1}{2}(t-1-z)}    \notag \\    
    = &  \left( \sum_{s=0}^{t-1} \mathbb{E}\| \gamma_{s}G^{\left[ j\right]}({X}_{s};\xi_{s})\| (q + p \rho)^{\frac{1}{2}(t-s-1)} \right)^2.
\end{align}
Substituting Eq. (\ref{HSHZ}) into (\ref{xtxbar}), we have
\begin{align}\label{consensus2}
    & \sum_{i=1}^{n}\mathbb{E}\| \mathbf{x}^{(i)}_t  - \overline{{X}_{t}} \|^2 \leq 2\sum_{j=1}^{N} \|\mathbf{x}^{\left[j\right]}_0 - \mathbf{x}^{\left[ j\right]}_0 \frac{\mathbf{1}_n {\mathbf{1}_n}^\top}{n} \|^2 (q + p \rho^2)^t\notag \\
    & + 2\sum_{j=1}^{N}\left( \sum_{s=0}^{t-1} \mathbb{E}\| \gamma_{s}G^{\left[ j\right]}({X}_{s};\xi_{s})\| (q + p \rho^2)^{\frac{1}{2}(t-s-1)} \right)^2.
\end{align}
Note that $\left(\sum_{s=0}^{t-1} \|\gamma_{s}G^{\left[ j\right]}({X}_{s};\xi_{s})\| (q + p \rho^2)^{\frac{1}{2}(t-s-1)} \right)^2$ has the same structure with the lemma of \cite{DCD-PSGD}, then summing Eq. (\ref{consensus2}) from $t=1$ to $t=T$, we have
\begin{align}\label{consensusequation}
    & \sum_{t=1}^{T}\sum_{i=1}^{n}\mathbb{E}\| \mathbf{x}^{(i)}_t  - \overline{{X}_{t}} \|^2 \notag \\
     \leq &  D_2 \sum_{j=1}^{N} \|\mathbf{x}^{\left[ j\right]}_0 - \mathbf{x}^{\left[ j\right]}_0 \frac{\mathbf{1}_n {\mathbf{1}_n}^\top}{n} \|^2 + D_1\sum_{j=1}^{N}  \sum_{t=1}^{T}\mathbb{E}\| \gamma_{s}G^{\left[ j\right]}({X}_{s};\xi_{s})\|^2  \notag \\
     \leq& D_2 \| X_0 - \overline{{X}_{0}} {\mathbf{1}_n}^\top \|^2_F  + D_1 \sum_{t=1}^{T} \gamma_{t}^2 \mathbb{E}\|G(X_t;\xi_{t}) \|^2_F,
\end{align}
where $D_1 = \frac{2}{(1 - (q+p\rho)^{\frac{1}{2}})^2} $ and $D_2 = \frac{2}{1-(q+p\rho^2)}$. 
\end{proof}
\end{theorem}
Note that if all workers have the same initial parameters,  $\| X_0 - \overline{{X}_{0}} {\mathbf{1}_n}^\top \|_F^2=0$, which means that the consensus is only affected by the stochastic gradients.

\subsection{Convergence Analysis}
Now we prove the convergence of SAPS-PSGD.

\begin{lemma}\label{lemma:gradientdifference}
Remove the $\frac{L}{2} \mathbb{E}\| \overline{Q_t}\|^2$ in Lemma 8 of \cite{DCD-PSGD} and rearrange it, we can have
\begin{align}\label{gradientdifference2}
    & \mathbb{E} \| \nabla f(\overline{X}_t) \|^2 + \left( 1 - L\gamma_t \right)\mathbb{E}\| \overline{\nabla f}(X_t)  \|^2 \notag \\
    \leq&  \frac{2}{\gamma_t}\left( \mathbb{E}f(\overline{X}_t)-f^* - (\mathbb{E}f(\overline{X}_{t+1}) -f^* )\right) \notag \\
    & + \frac{L^2}{n}\sum_{i=1}^{n}\mathbb{E}\| \mathbf{x}^{(i)}_t  - \overline{{X}_{t}} \|^2 + \frac{L\gamma_t \sigma^2}{n}.
\end{align}
\end{lemma}



\begin{lemma}\label{lemma:consensuswithfx}
Under the assumptions defined in Section \ref{subsec:assumptions}, if $X_t$ is iteratively updated by Eq. (\ref{iterationXt}), then we have
\begin{multline*}
    \sum_{t=1}^{T}(1-3D_1 L^2 \gamma_t^2)\sum_{i=1}^{n}\mathbb{E}\|\mathbf{x}^{(i)}_t  - \overline{{X}_{t}}  \|^2  \leq  D_1 n (\sigma^2 + 3 \zeta^2)\sum_{t=1}^{T}\gamma_t^2   \notag \\
     + 3n D_1 \sum_{t=1}^{T}\gamma_t^2 \| \nabla f(\overline{X}_{t})\|^2 + D_2 \| X_0 - \overline{{X}_{0}} {\mathbf{1}_n}^\top \|^2_F.
\end{multline*}
\begin{proof}
Combining (\ref{consensusequation}) and Lemma 12 in \cite{DCD-PSGD}, we have
\begin{align*}
    & \sum_{t=1}^{T}\sum_{i=1}^{n}\mathbb{E}\| \mathbf{x}^{(i)}_t  - \overline{{X}_{t}} \|^2 \notag \\
    \leq &  D_2 \| X_0 - \overline{{X}_{0}} {\mathbf{1}_n}^\top \|^2_F + D_1 \sum_{t=1}^{T} \gamma_{t}^2 \|G(X_t;\xi_{t}) \|^2_F \notag \\
    \leq &  D_2 \| X_0 - \overline{{X}_{0}} {\mathbf{1}_n}^\top \|^2_F + D_1 n(\sigma^2+3\zeta^2) \sum_{t=1}^{T} \gamma_{t}^2 + \notag\\
    & 3D_1 L^2\sum_{t=1}^{T}\gamma_{t}^2\sum_{i=1}^{n}\|\mathbf{x}^{(i)}_t  - \overline{{X}_{t}}  \|^2  +  3D_1 n\sum_{t=1}^{T}\gamma_{t}^2\mathbb{E}\| \nabla f(\overline{X}_{t}) \|^2 .\notag 
\end{align*}
By rearranging it, we obtain
\begin{multline*}
\sum_{t=1}^{T}(1-3D_1 L^2 \gamma_t^2)\sum_{i=1}^{n}\mathbb{E}\|\mathbf{x}^{(i)}_t  - \overline{{X}_{t}}  \|^2  \leq   D_1 n (\sigma^2 + 3 \zeta^2)\sum_{t=1}^{T}\gamma_t^2   \notag \\
    + 3n D_1 \sum_{t=1}^{T}\gamma_t^2 \| \nabla f(\overline{X}_{t})\|^2 + D_2 \| X_0 - \overline{{X}_{0}} {\mathbf{1}_n}^\top \|^2_F.
\end{multline*}
If $1-3D_1 L^2 \gamma_t^2>0 $, then $\sum_{t=1}^{T}\sum_{i=1}^{n}\mathbb{E}\|\mathbf{x}^{(i)}_t  - \overline{{X}_{t}}  \|^2 $ is bounded.
\end{proof}
\end{lemma}

\begin{theorem}\label{theorem:converge}
Under the assumptions in Section \ref{subsec:assumptions}, if $ \gamma_t =\gamma$ and $1-3D_1 L^2 \gamma >0$ for SAPS-PSGD, then
\begin{align}\label{equ:theo}
    & \frac{1}{T}\sum_{t=1}^{T}\mathbb{E} \| \nabla f(\overline{X}_t) \|^2 \notag \\
    \leq&  \frac{6\sigma(f(X_0) - f^*) + 3\sigma}{2\sqrt{nT}} + \frac{6\sqrt{3}L(f(X_0)-f^*)+2L^2 D_1 n}{T} \notag \\
    & + \frac{3L^2 D_1 n \zeta^2}{\sigma^2 T} + \frac{2L^2 D_2 \| X_0 - \overline{{X}_{0}} {\mathbf{1}_n}^\top \|^2_F}{nT},
\end{align}
where $f^*$ is the optimal solution.
\begin{proof}
According to Lemma \ref{lemma:consensuswithfx}, if we fix $\gamma_t$ to satisfy $1-3D_1 L^2 \gamma >0$ and sum both sides of (\ref{gradientdifference2}), we obtain
\begin{align*}
    & \sum_{t=1}^{T}\mathbb{E} \| \nabla f(\overline{X}_t) \|^2 + \sum_{t=1}^{T} \left( 1 - L\gamma \right)\mathbb{E}\| \overline{\nabla f}(X_t)  \|^2 \notag \\
    \leq &  \frac{2}{\gamma}\left( f(X_0) -f^* \right) + \frac{L^2}{n} ( \frac{D_1 n(\sigma^2 + 3 \zeta^2)T\gamma^2}{1-3D_1 L^2 \gamma^2} \notag \\
    & + \frac{3n D_1 \gamma^2}{1-3D_1 L^2 \gamma^2}\sum_{t=1}^{T}\| \nabla f(\overline{X}_{t})\|^2 + \frac{D_2 \| X_0 - \overline{{X}_{0}} {\mathbf{1}_n}^\top \|^2_F}{1-3D_1 L^2 \gamma^2} ) \notag \\
    & + \frac{LT\gamma \sigma^2}{n} \notag.
\end{align*}
Then we have
\begin{align}
    & \frac{1 - 6 D_1 L^2 \gamma^2}{1-3D_1 L^2 \gamma^2} \sum_{t=1}^{T}\mathbb{E} \| \nabla f(\overline{X}_t) \|^2 + \sum_{t=1}^{T} \left( 1 - L\gamma \right)\mathbb{E}\| \overline{\nabla f}(X_t)  \|^2 \notag \\
    & \leq  \frac{2}{\gamma}( f(X_0) -f^* ) + ( \frac{L^2 D_1 T \gamma^2}{1-3D_1 L^2 \gamma^2} + \frac{L\gamma T}{n} )\sigma^2 \notag \\
    & + \frac{3L^2 D_1 T \gamma^2 \zeta^2}{1-3D_1 L^2 \gamma^2} + \frac{L^2 D_2 \| X_0 - \overline{{X}_{0}} {\mathbf{1}_n}^\top \|^2_F}{n(1-3D_1 L^2 \gamma^2)}.
\end{align}
Setting $\gamma = \frac{1}{2\sqrt{3D_1}L+\frac{\sigma}{\sqrt{n}}\sqrt{T}}$, it yields
\begin{align*}
    3 D_1 L^2 \gamma^2  \leq \frac{1}{4} , \frac{1 - 6 D_1 L^2 \gamma^2}{1-3D_1 L^2 \gamma^2}  \ge \frac{2}{3}, 
    1 - L\gamma  > 0.
\end{align*}
By removing the $\mathbb{E}\| \overline{\nabla f}(X_t)  \|^2 $ on the LHS and substituting $\frac{1 - 6 D_1 L^2 \gamma^2}{1-3D_1 L^2 \gamma^2}$ with $\frac{2}{3}$, we can have the form of (\ref{equ:theo}), which concludes the proof.
\end{proof}
\textbf{Remark.} The theorem indicates that SAPS-PSGD has a convergence rate of $O(\frac{1}{\sqrt{n T}})$, and the sparsity dose not sacrifice the convergence performance when $T$ is large enough. The convergence rate is consistent to D-PSGD \cite{lian2017can} and the original PSGD \cite{dekel2012optimal}.
\end{theorem}

\section{Experimental Study} \label{sec:evaluation}

In this section, we compare the performance of SAPS-PSGD with PSGD (with all-reduce), TopK-PSGD \cite{lin2017deep,renggli2019sparcml}, FedAvg \cite{mcmahan2016communication}, Sparse FedAvg (S-FedAvg)\cite{konevcny2016federated}, D-PSGD \cite{lian2017can}, and DCD-PSGD \cite{DCD-PSGD}. First, we evaluate the convergence performance with respect to the number of iterations on 32 workers without considering the network bandwidth. Second, we compare the generated communication traffic during the training to achieve a target validation accuracy. Third, we evaluate the bandwidth utilization of SAPS-SGD under two emulated distributed environments: one with 14 workers located at 14 cities in Fig. \ref{fig:networkspeed}, and another one with 32 workers with randomly generated communication speed between any two workers.  

\subsection{Experimental Settings}
\begin{table}[!ht]
	\centering
	\caption{Experimental settings}
	\label{table:hyperparam}
	\begin{tabular}{|l|l|l|l|l|}
		\hline
	Model	 &   \# Params & Batch Size & LR & \# Epochs \\\hline\hline
	MNIST-CNN & 6,653,628 & 50 & 0.05 &100\\\hline
	CIFAR10-CNN	& 7,025,886 & 100 & 0.04  &320\\\hline
	ResNet-20	& 269,722 & 64 & 0.1  &160\\\hline
	\end{tabular}
\end{table}
We use two commonly used data sets for performance evaluation: MNIST \cite{lecun2010mnist} with 60,000 training images and 10,000 validation images in 10 classes, and CIFAR10 \cite{krizhevsky2010cifar} with 50,000 training images and 10,000 validation images in 10 classes. Regarding the models, we use several representative convolutional neural networks (CNNs) to verify the convergence and communication cost: 1) The same CNN model (MNIST-CNN) with \cite{mcmahan2016communication} training on the MNIST data set. 2) The same CNN model (CIFAR10-CNN) with \cite{mcmahan2016communication} training on the CIFAR10 data set. 3) ResNet-20 \cite{he2016deep} with skip connections training on the CIFAR10 data set. The experimental settings are shown in Table \ref{table:hyperparam}. 

For FedAvg and S-FedAvg algorithms, we set the ratio of chosen workers to 0.5 and set a compression ratio $c=100$ for S-FegAvg, which are the same as \cite{mcmahan2016communication}. For TopK-PSGD, we set $c=1000$, which is the same as \cite{lin2017deep}. For DCD-PSGD, we set $c=4$ (the same as \cite{DCD-PSGD}) to achieve a good convergence. Note that in DCD-PSGD, if $c$ is larger than 4, it would lose much accuracy; and if $c$ is set to $100$ or $1000$ like S-FegAvg or TopK-PSGD, it would not converge at all. For our SAPS-PSGD, we set $c=100$.

\subsection{Convergence Performance}
\begin{figure*}[!ht]
	\centering
	\subfigure[MNIST-CNN]
	{
		\includegraphics[width=0.3\linewidth]{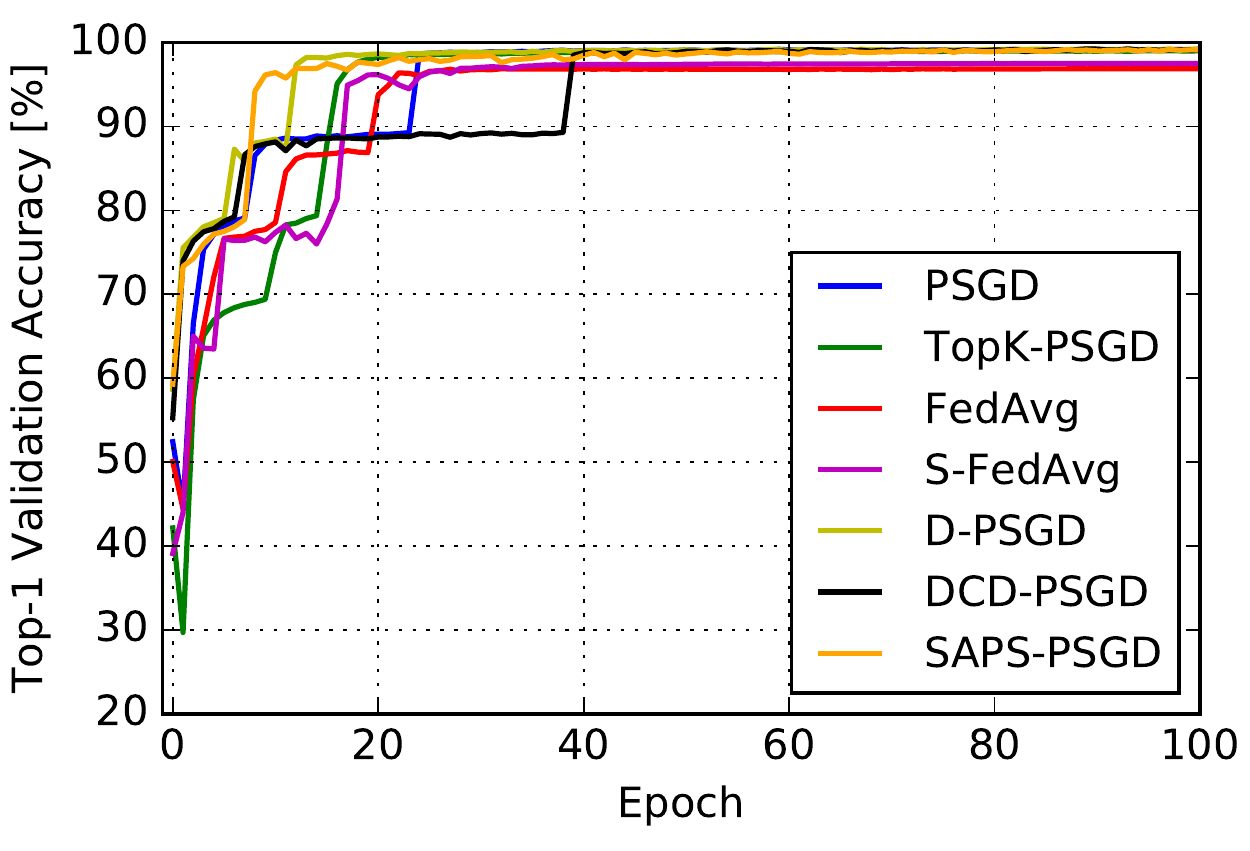}
	}
	\subfigure[CIFAR10-CNN]
	{
		\includegraphics[width=0.3\linewidth]{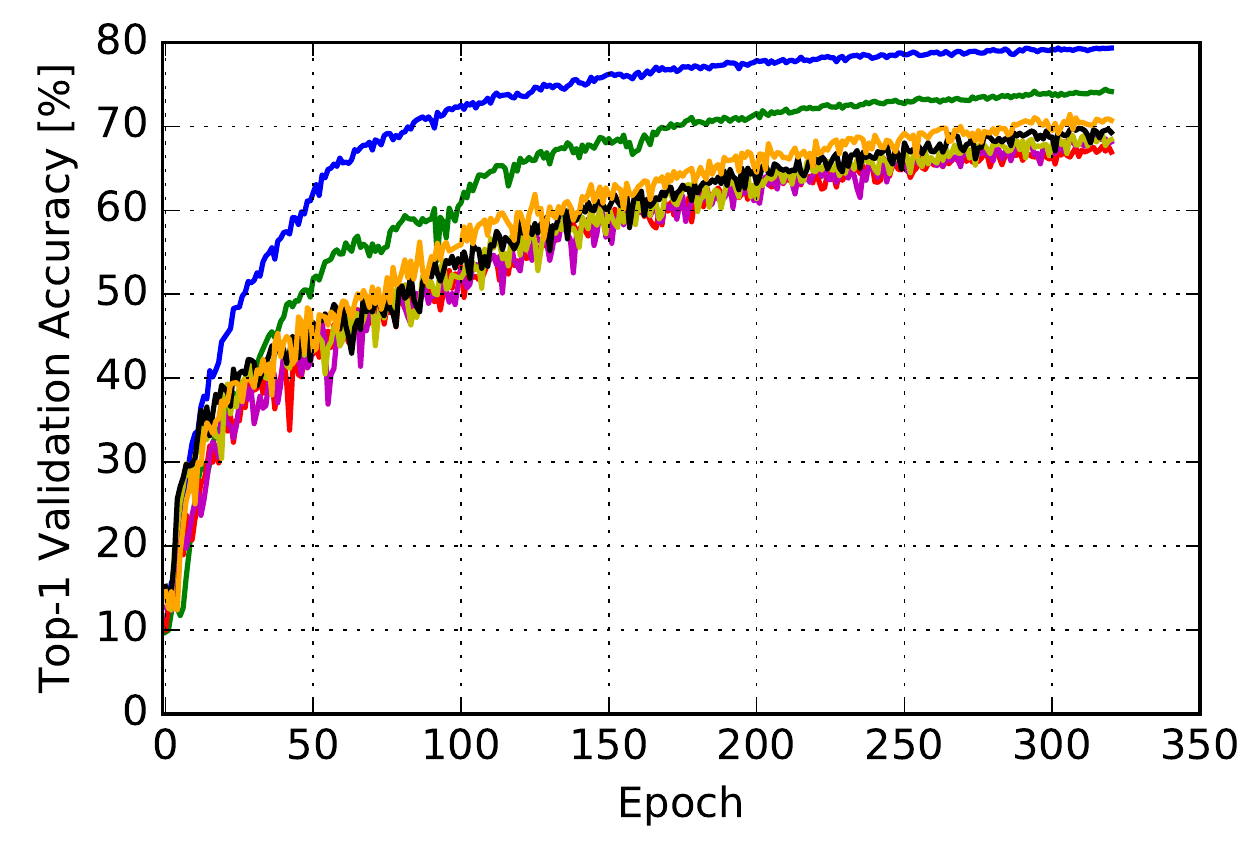}
	}
	\subfigure[ResNet-20]
	{
		\includegraphics[width=0.3\linewidth]{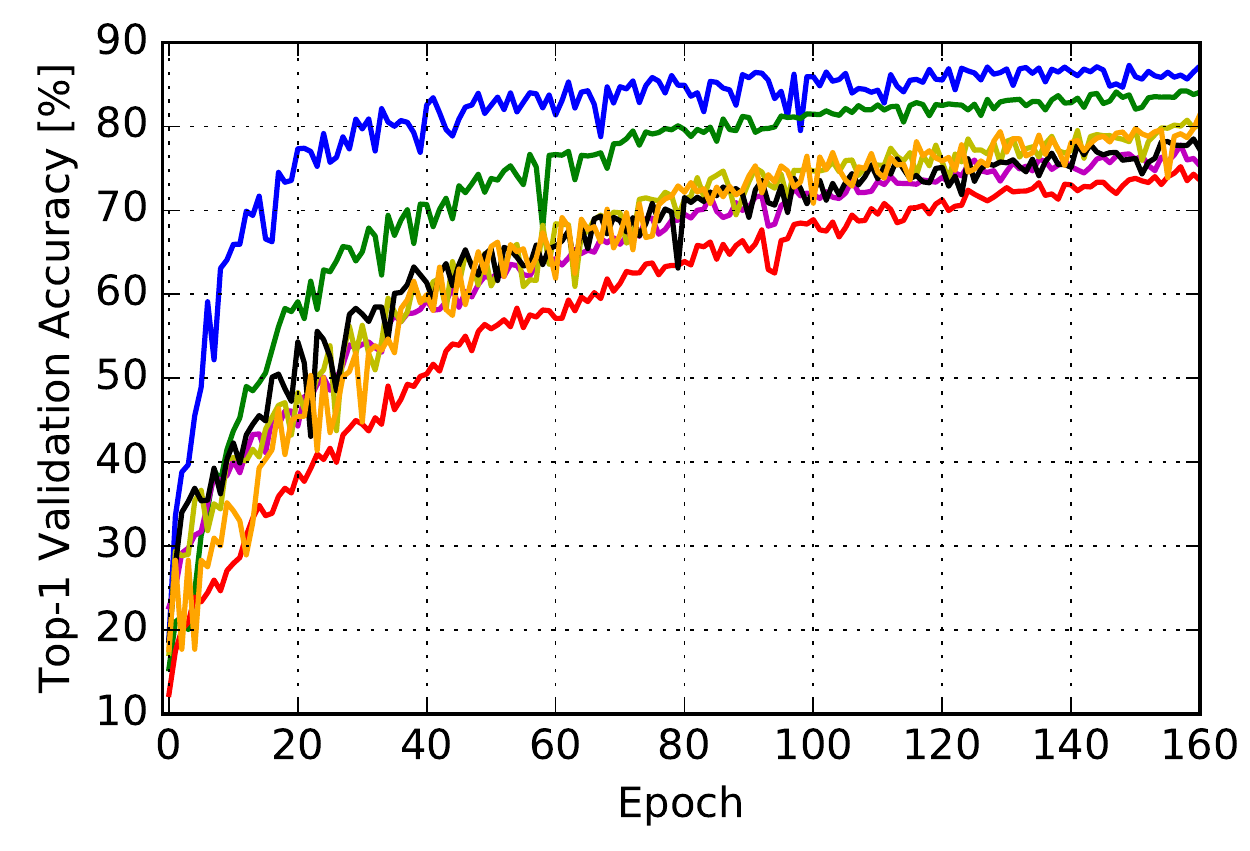}
	}
	\caption{Convergence performance of different algorithms on three models with 32 workers.}
	\vspace{-10pt}
	\label{fig:accvsepoch}
\end{figure*}
The top-1 validation accuracy with respect to the number of epochs is shown in Fig. \ref{fig:accvsepoch}; and the final model accuracy is shown in Table \ref{table:acc}. We can see that SAPS-PSGD achieves similar convergence performance with D-PSGD and finally obtains higher validation accuracy than other algorithms except PSGD and TopK-PSGD. The good convergence performance verifies that SAPS-PSGD has comparable convergence rate with D-PSGD. Compared to PSGD and TopK-PSGD, it is noticed that SAPS-PSGD has some accuracy loss on CIFAR10 under the same number of training epochs. The main reason is that in SAPS-PSGD, each worker only communicates with a single peer, and it requires some iterations to achieve the consensus as shown in Section \ref{subsec:consensus} while PSGD and TopK-PSGD are with the fully connected structure (each worker receives all other workers' information at each communication round) so that they have the consensus at every iteration. In contrast, PSGD and TopK-PSGD require high communications, which is demonstrated in the following section.
\begin{table}[!h]
    \centering
    \caption{Comparison of top-1 validation accuracy (in percentage).}\label{table:acc}
    \begin{tabular}{|c|c|c|c|}
    \hline
       Algorithm  & MNIST-CNN & CIFAR10-CNN  & ResNet-20  \\\hline\hline
       PSGD  & 99.19 & 79.35 & 87.27 \\\hline
       TopK-PSGD & 99.03 & 74.43 & 84.23 \\\hline
       FedAvg  & 96.88 & 67.66 & 75.2 \\\hline
       S-FedAvg & 97.51 & 68.8 & 77.78 \\\hline
       D-PSGD  & 99.24 & 69.07 & 80.74 \\\hline
       DCD-PSGD & 99.24 & 69.78 & 78.51 \\\hline
       SAPS-PSGD (ours) & 99.17 & 71.44 & 81.42 \\\hline
    \end{tabular}
\end{table}

\subsection{Convergence vs. Communication Cost}
\begin{figure*}[!ht]
	\centering
	\subfigure[MNIST-CNN]
	{
	\includegraphics[width=0.3\linewidth]{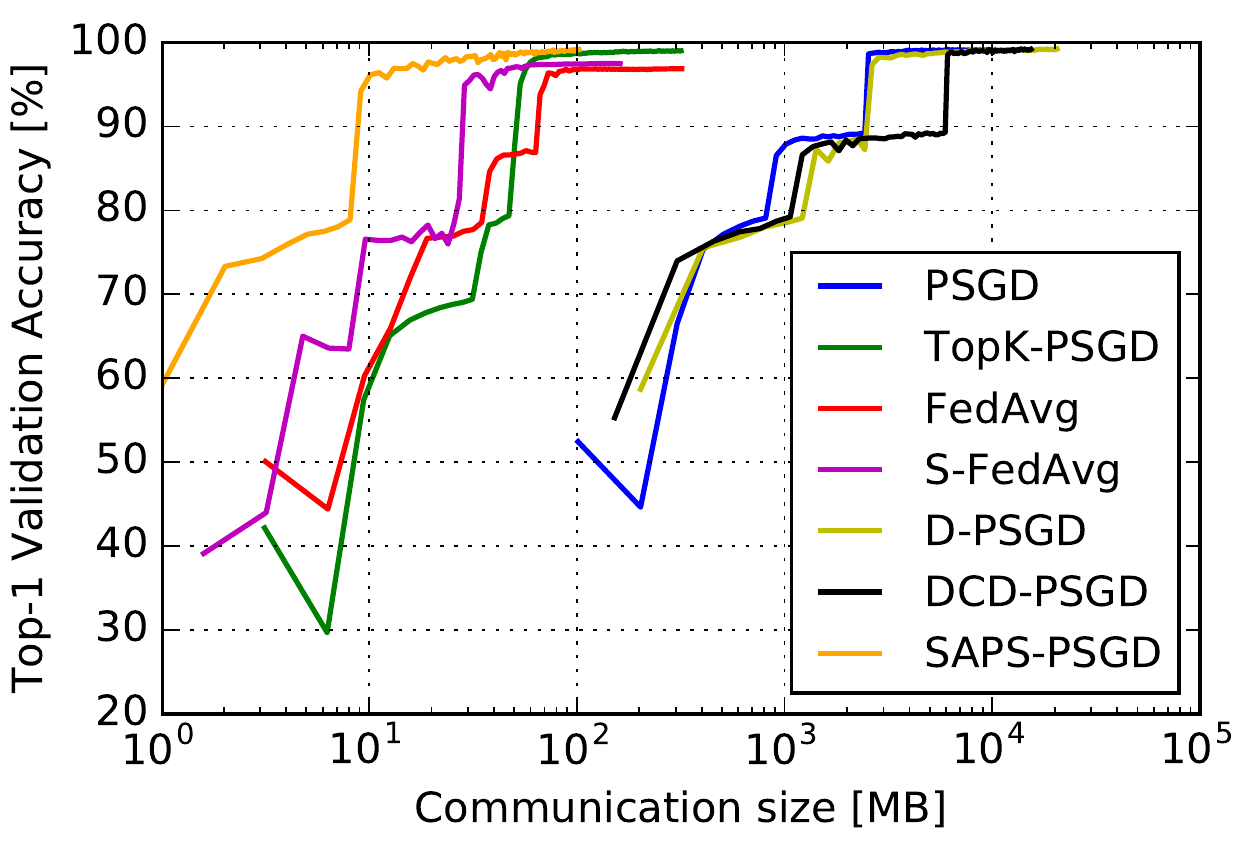}
	}
	\subfigure[CIFAR10-CNN]
	{
	\includegraphics[width=0.3\linewidth]{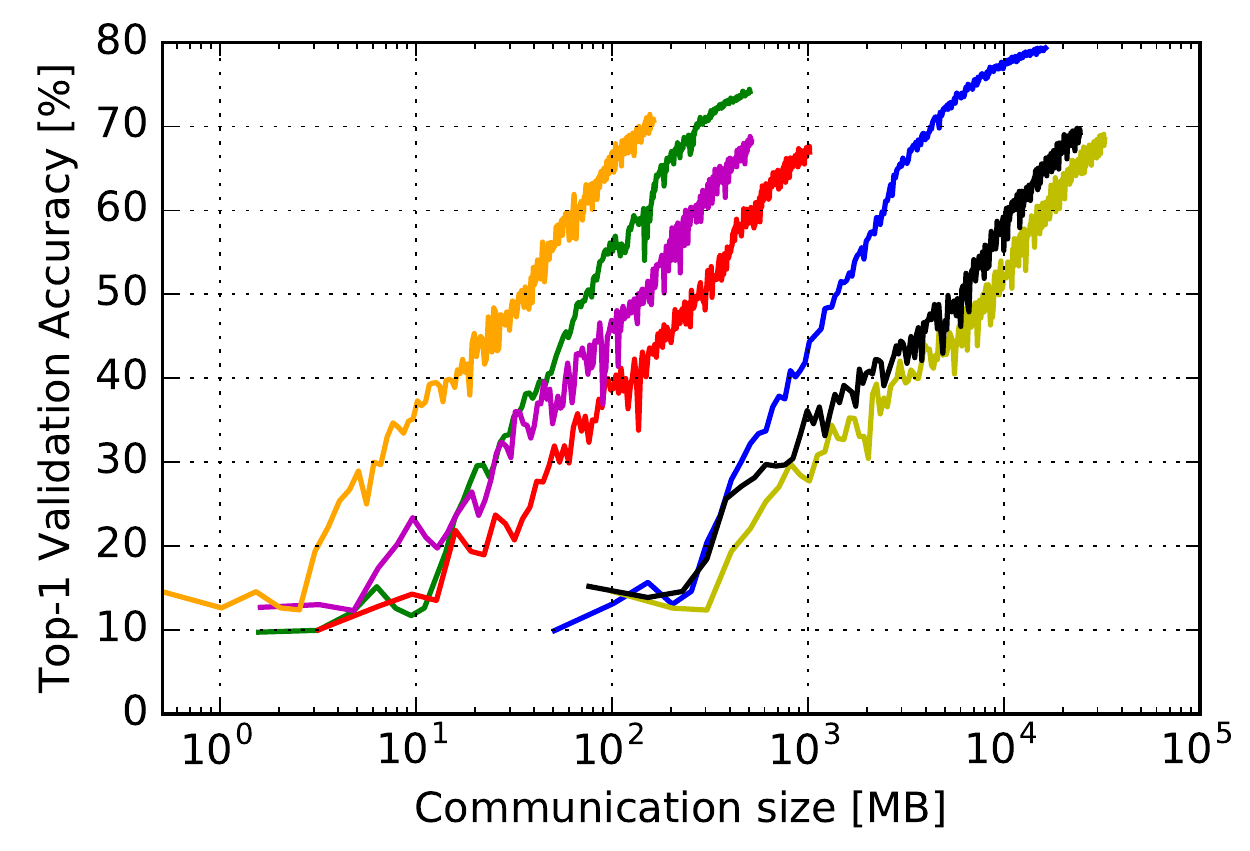}
	}
	\subfigure[ResNet-20]
	{
	\includegraphics[width=0.3\linewidth]{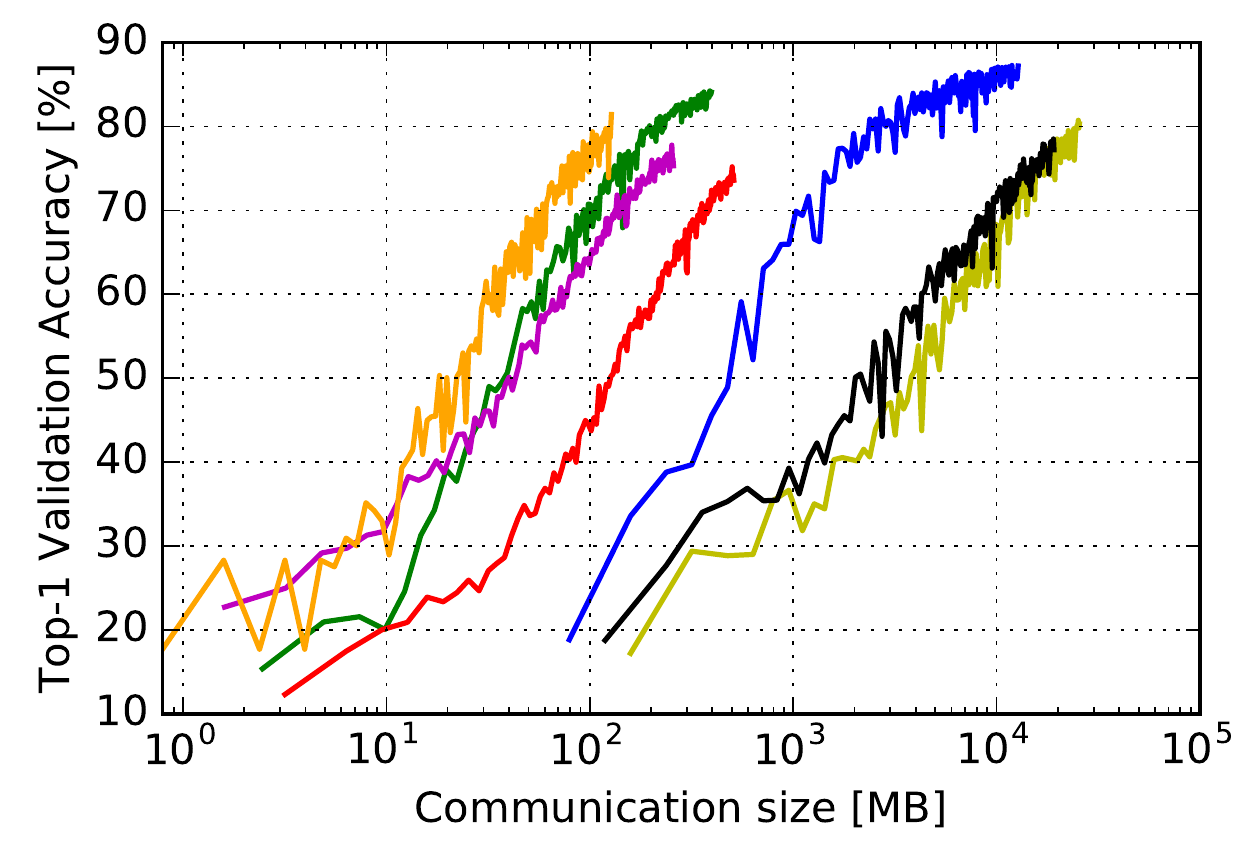}
	}
	\caption{The validation accuracy with respect to the communication size on a training worker under the 32-worker distributed setting.}
	\label{fig:accvscomm}
\end{figure*}

The comparison of the validation accuracy vs. communication size on three evaluated models is shown in Fig. \ref{fig:accvscomm}. The experimental results show that our proposed SAPS-PSGD spends the smallest amount of communication to achieve the same level of accuracy. The results show that our SAPS-PSGD can sparsify a large number of parameters during the training to significantly reduce the communication cost while preserving the model accuracy. The sparsification and single-peer scheme of SAPS-PSGD result in a much lower communication traffic than existing ones.

On MNIST-CNN as shown in Fig. \ref{fig:accvscomm}(a), to achieve 96.8\% validation accuracy, FedAvg and S-FedAvg require about $70$MB and $32$MB accumulated communication traffic, and D-PSGD and DCD-PSGD also require about $1800$MB and $5000$MB, while our SAPS-PSGD only requires $10$MB accumulated communication traffic which is around $7\times$ and $4.5\times$ smaller than FedAvg and S-FedAvg respectively. However, D-PSGD requires a worker to send its full model to it peers, which is very communication-intensive. On CIFAR10-CNN as shown in Fig. \ref{fig:accvscomm}(b), at achieve 67\% accuracy, SAPS-PSGD spends about $120$MB which is $8.3\times$ and $4.2\times$ smaller than FedAvg ($1000$ MB) and S-FedAvg ($500$ MB) respectively. Again, D-PSGD suffers from the high communication traffic. On ResNet-20 as shown in Fig. \ref{fig:accvscomm}(c), to achieve 75\% accuracy, SAPS-PSGD spends more than $2\times$ and $3.1\times$ smaller communication size than the centralized algorithms and the decentralized algorithms, respectively.

\subsection{Bandwidth Utilization}
\begin{figure}[!ht]
	\centering
	\hspace{-15pt}
	\subfigure[14-worker]
	{
	\includegraphics[width=0.49\linewidth]{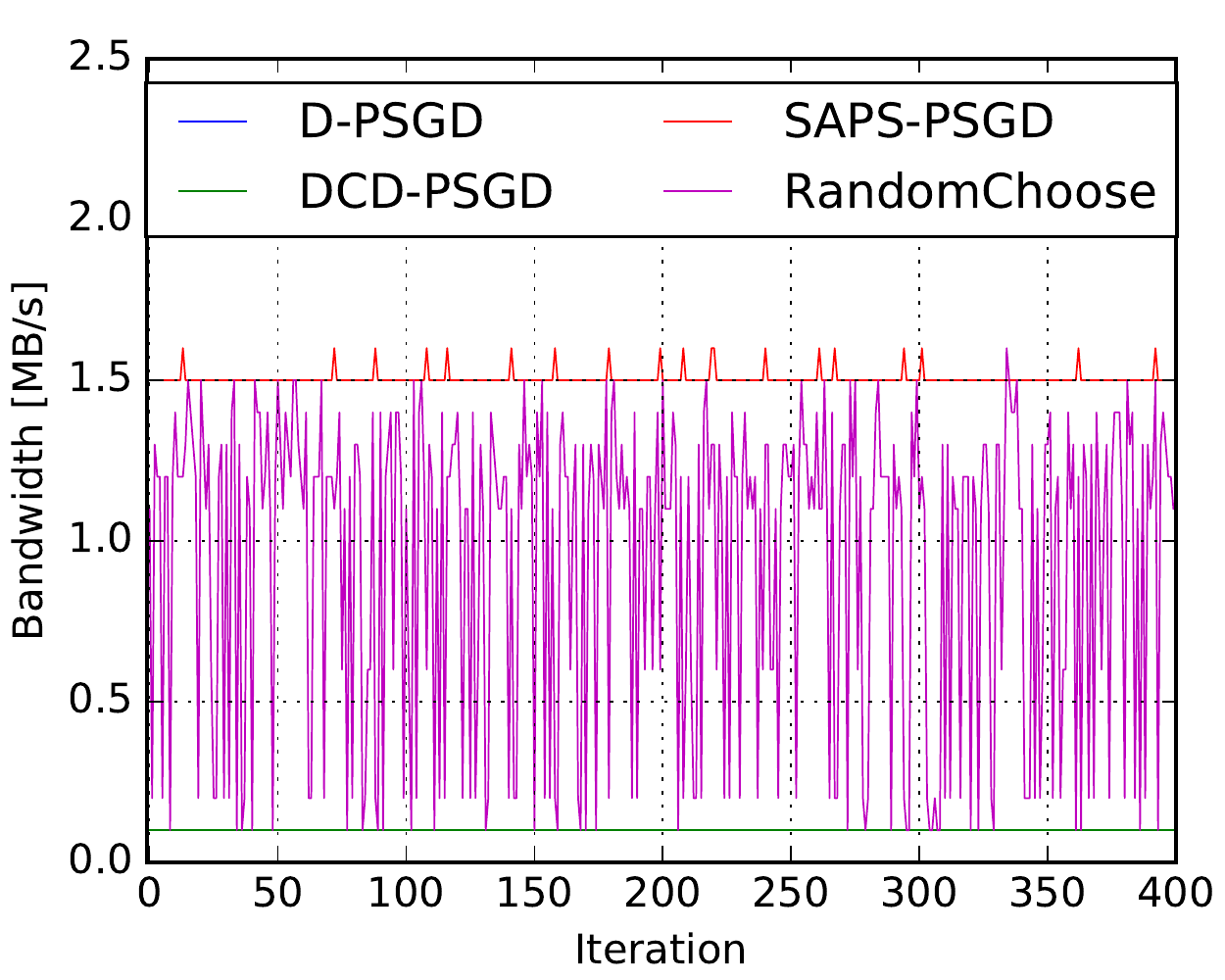}
	}\hspace{-10pt}
	\subfigure[32-worker]
	{
	\includegraphics[width=0.49\linewidth]{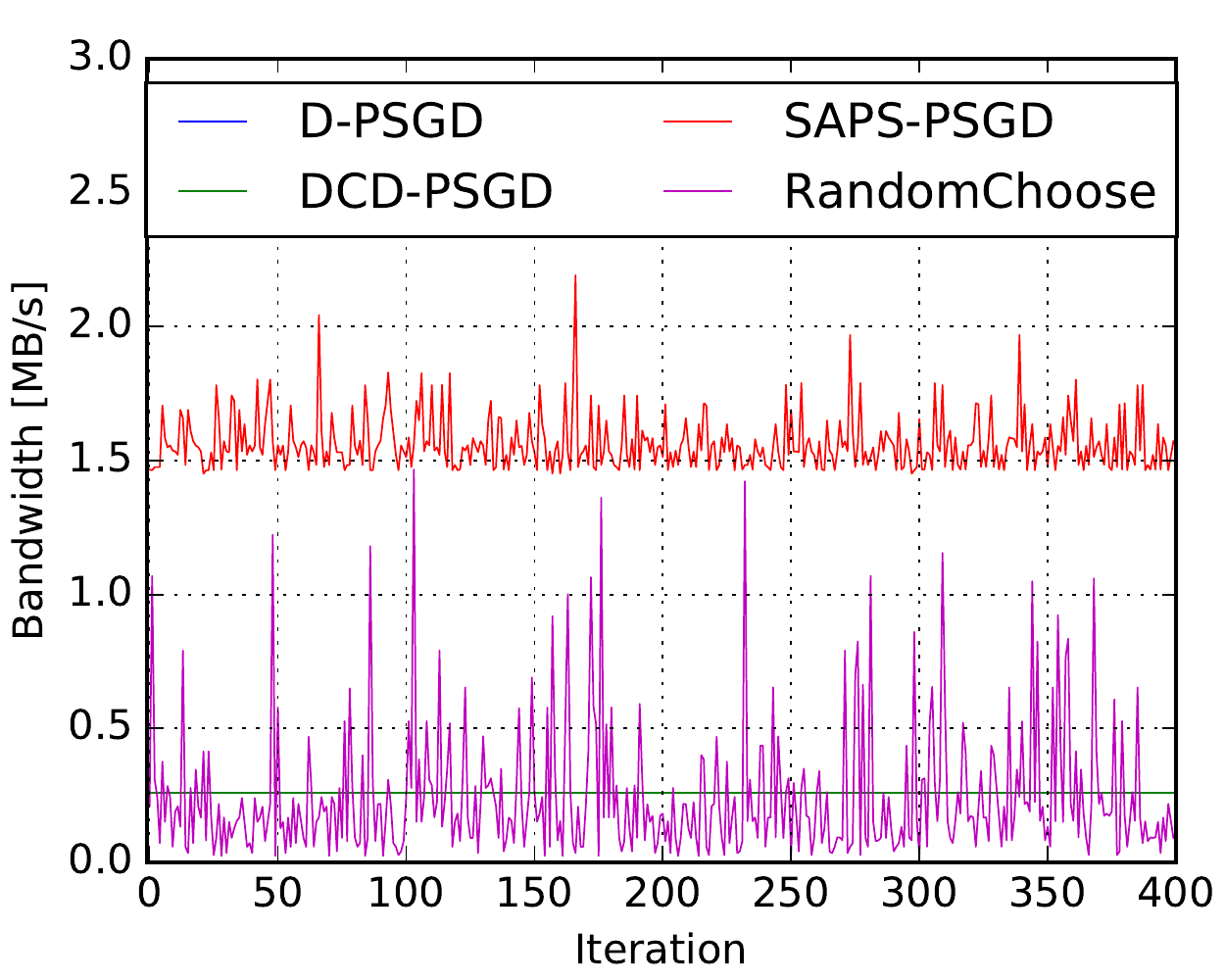}
	}
	\hspace{-10pt}
	\caption{Bandwidth utilization under two distributed environments.}
	\vspace{-10pt}
	\label{fig:bandwidthutil}
\end{figure}

We demonstrate the bandwidth utilization on a 14-worker environment (the bandwidths between two workers are simulated from Fig. \ref{fig:networkspeed}) and a 32-worker environment (the bandwidths between two workers are generated randomly from the range (0MB/s, 5MB/s] in a uniform distribution. The bandwidth utilization of different algorithms are shown in Fig. \ref{fig:bandwidthutil}. Note that FedAvg and S-FedAvg are centralized algorithms, in which the bandwidth utilization is only determined by the workers and the server but not related to the bandwidths between workers, so we exclude these algorithms from comparison. As mentioned in Section \ref{subsec:gossipmatrix}, finding a best bandwidth cycle in a general graph is very difficult. So here we randomly generate 5,000 different bandwidth matrices, and for every matrix we set a ring topology according to the order ${1\to 2 \to...\to 32 \to 1}$, avoiding the huge variance of different random matrices. Then we set the average value as our final bandwidth of D-PSGD and DCD-PSGD (ring-topology). Another way to choose the communication peers for decentralized training is to randomly do maximum match, which can reach consensus faster due to the same possibility of being chosen; but it suffers from low bandwidth. As shown in Fig. \ref{fig:bandwidthutil}, we only display the bandwidth utilization at the first 400 iterations for better visualization, and the remaining iterations have a similar pattern. The results show that SAPS-PSGD generally selects peers with higher bandwidth than D-PSGD (DCD-PSGD, PSGD and TopK-PSGD are also the same) and the random one. Random maximum match can gain better bandwidth than the ring topology because the expectation of lowest value of 16 random edges is higher than the ring topology with 32 edges.

\begin{figure*}[!ht]
	\centering
	\subfigure[MNIST-CNN]
	{
	\includegraphics[width=0.3\linewidth]{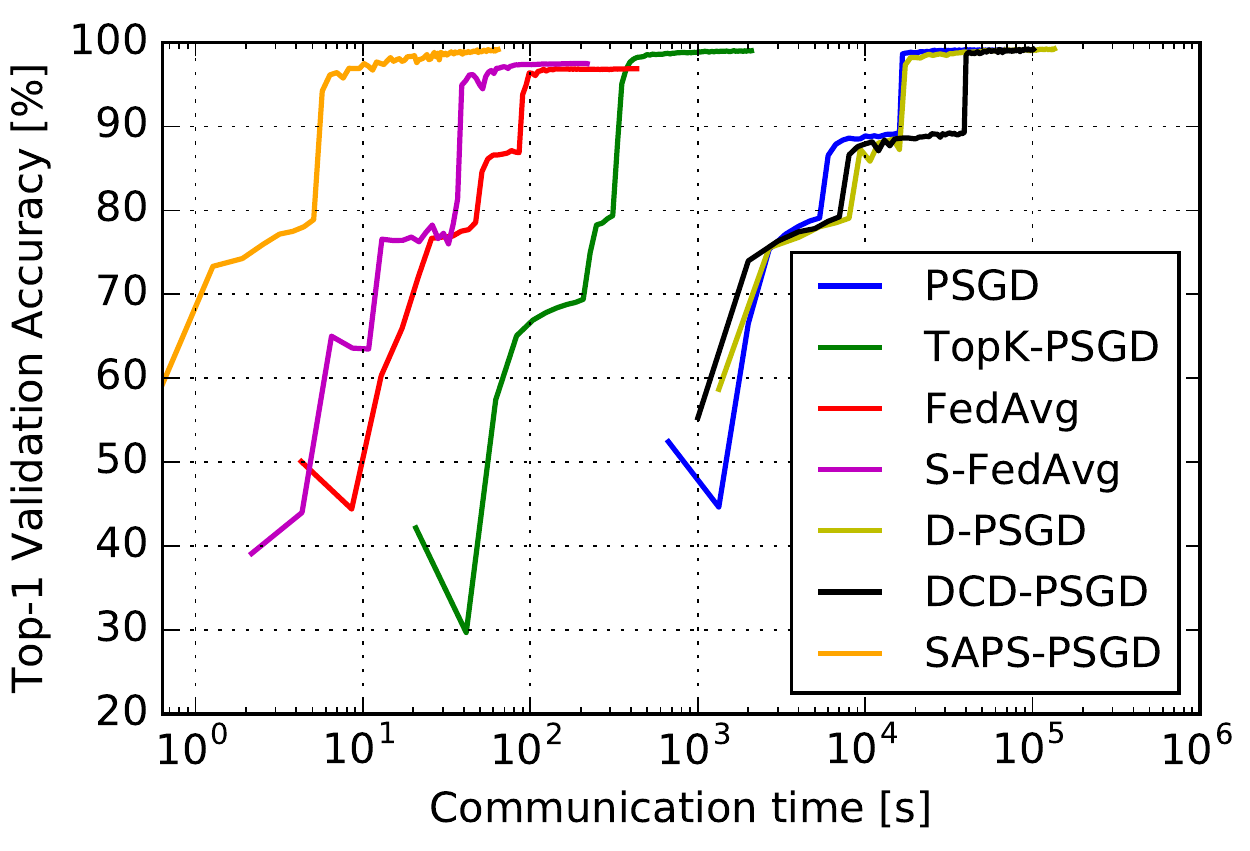}
	}
	\subfigure[CIFAR10-CNN]
	{
	\includegraphics[width=0.3\linewidth]{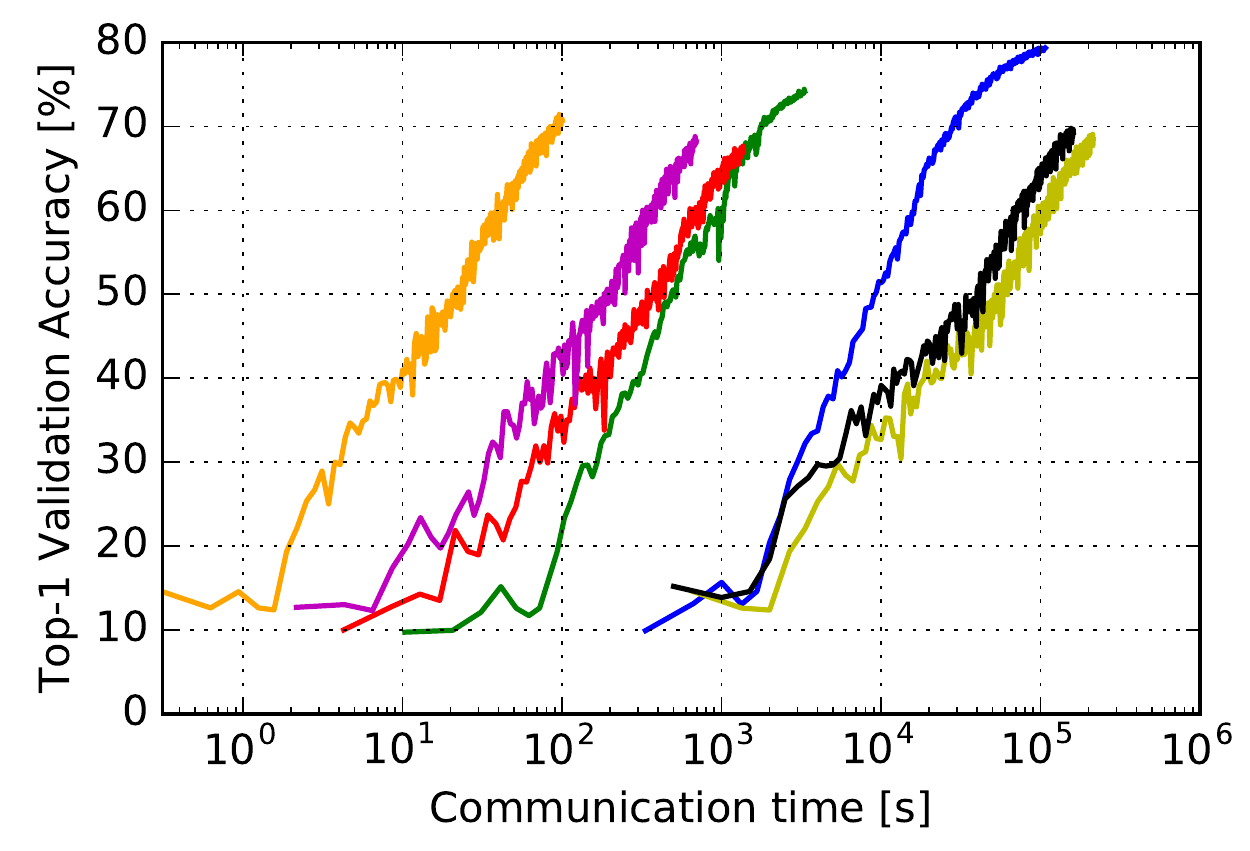}
	}
	\subfigure[ResNet-20]
	{
	\includegraphics[width=0.3\linewidth]{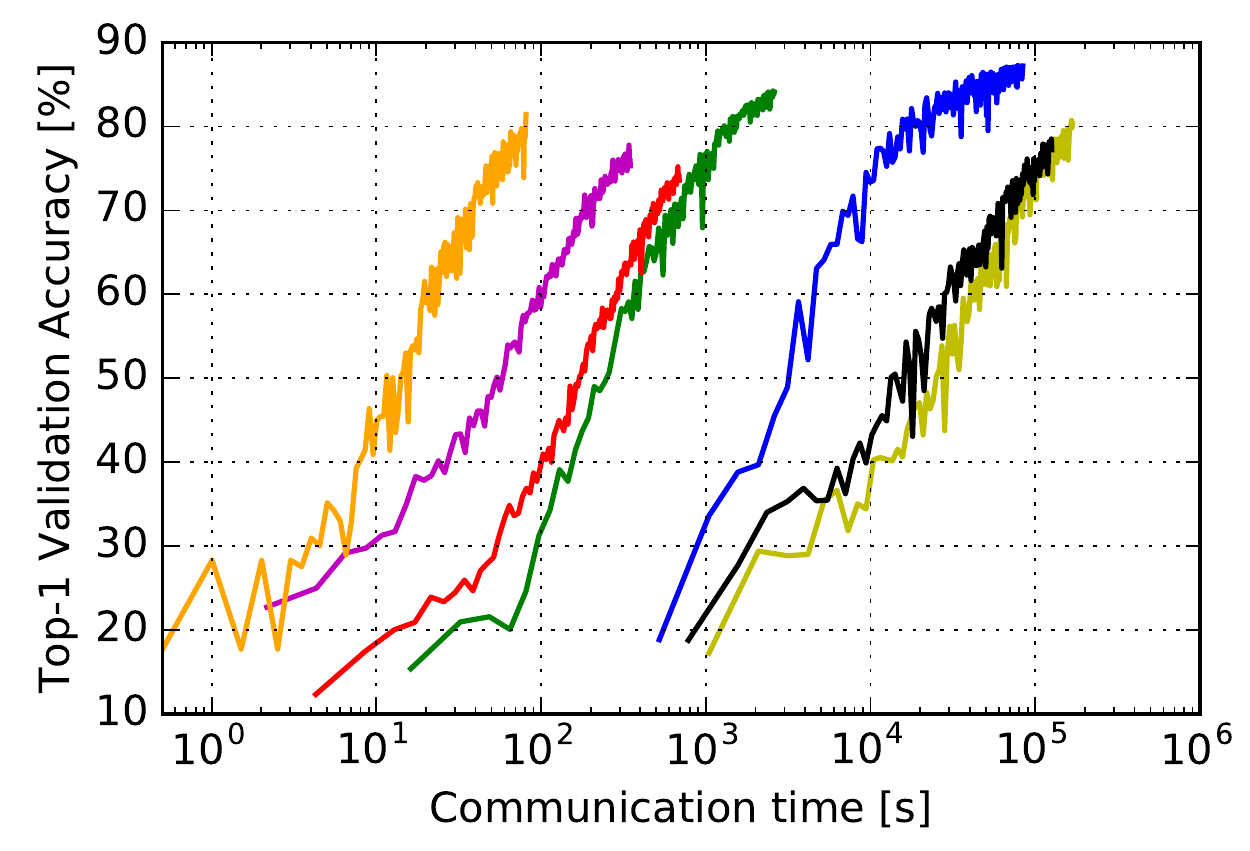}
	}
	\caption{The validation accuracy with respect to the communication time with randomly generated bandwidths for 32 workers.}
	\vspace{-8pt}
	\label{fig:accvstime}
\end{figure*}
Putting the communication traffic and the bandwidth together, we directly compare the model convergence vs. the communication time\footnote{Due to the diversity of computing resources (e.g., CPU and GPU), the computation time may be various. So we mainly focus on the comparison of communication time, while the end-to-end training time can also be obtained accordingly for the given specific processors.}, which is shown in Fig. \ref{fig:accvstime}. Here we calculate the bandwidths of FedAvg and S-FedAvg by choosing the server that has the maximum bandwidth. It can be seen that the improvement of SAPS-PSGD over other algorithms in communication time increases as we choose the pair of workers with higher bandwidth. To summarize, we compare all algorithms in both communication traffic and communication time to reach target accuracy considering the bandwidths between workers in Table \ref{table:overallcompare}. It is seen that the communication traffic is much lower than other algorithms due to the sparse and single-peer communication, and the improvement on the communication time is further boosted due to the adaptive peer selection to choose relatively high bandwidth communications.

\begin{table}[!h]
    \centering
    \caption{Communication traffic (MB) and time (second) at reaching target accuracy with bandwidth included in 32 workers.}\label{table:overallcompare}
    \addtolength{\tabcolsep}{-1.5pt}
    \begin{tabular}{|c|c|c||c|c||c|c|}
    \hline
         \multirow{3}{*}{Algorithm} & \multicolumn{2}{c||}{MNIST-CNN} &
         \multicolumn{2}{c||}{CIFAR10-CNN} & \multicolumn{2}{c|}{ResNet-20} \\
         & \multicolumn{2}{c||}{(96\%)} & \multicolumn{2}{c||}{(67\%)} & \multicolumn{2}{c|}{(75\%)} \\\cline{2-7}
         & Traffic & Time & Traffic & Time & Traffic & Time \\\hline\hline
        PSGD  & 2400 & 15800 & 3200 & 21000 & 1600 & 10500\\\hline
        TopK-PSGD & 55 & 360 & 210 & 1380 & 160 & 1000 \\\hline
        FedAvg & 70 & 94 & 1000 & 1350 & 510 & 680 \\ \hline
        S-FedAvg & 32 & 43 & 500 & 680 & 250 & 340 \\\hline
        D-PSGD  & 2600  & 17100 & 30000 & 197000 & 18000 & 118000 \\\hline
        DCD-PSGD & 6000 & 39000 & 22000 & 144000 & 15000 & 98000 \\\hline
        SAPS-PSGD & \textbf{10} & \textbf{6} & \textbf{120} & \textbf{75} & \textbf{80} & \textbf{50} \\\hline
    \end{tabular}
    \vspace{-10pt}
\end{table}

\section{Related Work} \label{sec:relatedwork}
The communication problem is common in distributed machine learning in data centers with relatively high bandwidth connections or in geo-distributed workers with low bandwidth and unstable connections. There exist various techniques addressing the communication problem in distributed learning.

\textbf{Gradient/Model Compression}. Gradient compression is a key technique to reduce the communication traffic by quantizing or sparsifing the gradients. Quantization \cite{seide20141,wen2017terngrad,hubara2017quantized,alistarh2017qsgd} reduces the number of bits to represent the data and can achieve up to 32$\times$ traffic reduction. Sparsification \cite{chen2017adacomp,lin2017deep,wangni2018gradient,zhao2018federated,shi2019aconvergence,shi2020layer} is orthogonal to quantization, and it can significantly reduce the communication traffic by zeroing out a large proportion of elements. Gaia \cite{hsieh2017gaia} is another form of gradient compression, who filters out ``insignificant'' updates. However, either the current sparsification method applied in the centralized architecture or the decentralized architecture makes little communication saving on the server side or the worker side.

\textbf{Decentralized Training}. To eliminate the bandwidth problem of the central server, some researchers propose the decentralized distributed training algorithms \cite{lian2017can,mcmahan2017communication,sirb2018decentralized} that training workers only need to communicate with their peers instead of the central server or other $n-1$ workers. As a result, the communication overheads are main in the model transferring among multiple workers. However, the size of models could be very large such that the communication bottleneck still exists. In \cite{DCD-PSGD}, the authors propose algorithms that only requires the worker to transfer a compressed model with convergence guarantees. And they theoretically provide a convergence analysis for this compression method.

\section{Conclusion} \label{sec:conclusion}
In this paper, we proposed a communication-efficient decentralized learning algorithm (SAPS-PSGD) with sparsification and adaptive peer selection. The decentralized architecture eliminates the bandwidth bottleneck at the server side, the sparsification technique significantly reduces the communication cost for workers which only need to exchange a highly sparse model with a single peer, and the adaptive peer selection feature can better utilize the bandwidths of workers. We provided a detailed analysis for the convergence property of SAPS-PSGD, which concludes that SAPS-PSGD has convergence guarantees and has a comparable convergence rate with traditional parallel SGD algorithms. Extensive experiments were conducted to verify the convergence performance, communication traffic reduction, bandwidth utilization and communication time of our algorithm compared to existing popular ones. Experimental results showed that SAPS-PSGD not only significantly reduces the communication traffic, but it also selects relatively high bandwidth peers for communication to increase the high network resource usage while preserving the convergence performance.

\bibliographystyle{IEEEtran}
\bibliography{cites-short}

\begin{thebibliography}{10}
\providecommand{\url}[1]{#1}
\csname url@samestyle\endcsname
\providecommand{\newblock}{\relax}
\providecommand{\bibinfo}[2]{#2}
\providecommand{\BIBentrySTDinterwordspacing}{\spaceskip=0pt\relax}
\providecommand{\BIBentryALTinterwordstretchfactor}{4}
\providecommand{\BIBentryALTinterwordspacing}{\spaceskip=\fontdimen2\font plus
\BIBentryALTinterwordstretchfactor\fontdimen3\font minus
  \fontdimen4\font\relax}
\providecommand{\BIBforeignlanguage}[2]{{%
\expandafter\ifx\csname l@#1\endcsname\relax
\typeout{** WARNING: IEEEtran.bst: No hyphenation pattern has been}%
\typeout{** loaded for the language `#1'. Using the pattern for}%
\typeout{** the default language instead.}%
\else
\language=\csname l@#1\endcsname
\fi
#2}}
\providecommand{\BIBdecl}{\relax}
\BIBdecl

\bibitem{dean2012large}
J.~Dean, G.~Corrado, R.~Monga, K.~Chen, M.~Devin, M.~Mao, A.~Senior, P.~Tucker,
  K.~Yang, Q.~V. Le \emph{et~al.}, ``Large scale distributed deep networks,''
  in \emph{NeurlPS}, 2012, pp. 1223--1231.

\bibitem{li2014scaling}
M.~Li, D.~G. Andersen, J.~W. Park, A.~J. Smola, A.~Ahmed, V.~Josifovski,
  J.~Long, E.~J. Shekita, and B.-Y. Su, ``Scaling distributed machine learning
  with the parameter server.'' in \emph{OSDI}, vol.~14, 2014, pp. 583--598.

\bibitem{li2014communication}
M.~Li, D.~G. Andersen, A.~J. Smola, and K.~Yu, ``Communication efficient
  distributed machine learning with the parameter server,'' in \emph{NeurlPS},
  2014, pp. 19--27.

\bibitem{zhou2017kunpeng}
J.~Zhou, X.~Li, P.~Zhao, C.~Chen, L.~Li, X.~Yang, Q.~Cui, J.~Yu, X.~Chen,
  Y.~Ding \emph{et~al.}, ``Kunpeng: Parameter server based distributed learning
  systems and its applications in alibaba and ant financial,'' in \emph{ACM
  SIGKDD}.\hskip 1em plus 0.5em minus 0.4em\relax ACM, 2017, pp. 1693--1702.

\bibitem{konevcny2016federated}
J.~Kone{\v{c}}n{\`y}, H.~B. McMahan, F.~X. Yu, P.~Richt{\'a}rik, A.~T. Suresh,
  and D.~Bacon, ``Federated learning: Strategies for improving communication
  efficiency,'' \emph{NeurlPS}, 2016.

\bibitem{mcmahan2017communication}
B.~McMahan, E.~Moore, D.~Ramage, S.~Hampson, and B.~A. y~Arcas,
  ``Communication-efficient learning of deep networks from decentralized
  data,'' in \emph{AISTATS}, 2017, pp. 1273--1282.

\bibitem{hsieh2017gaia}
K.~Hsieh, A.~Harlap, N.~Vijaykumar, D.~Konomis, G.~R. Ganger, P.~B. Gibbons,
  and O.~Mutlu, ``Gaia: Geo-distributed machine learning approaching
  $\{$LAN$\}$ speeds,'' in \emph{USENIX}, 2017, pp. 629--647.

\bibitem{shi2018performance}
S.~Shi, W.~Qiang, and X.~Chu, ``Performance modeling and evaluation of
  distributed deep learning frameworks on {GPUs},'' in \emph{DataCom}, 2018,
  pp. 949--957.

\bibitem{abadi2016tensorflow}
M.~Abadi, P.~Barham, J.~Chen, Z.~Chen, A.~Davis, J.~Dean, M.~Devin,
  S.~Ghemawat, G.~Irving, M.~Isard \emph{et~al.}, ``Tensorflow: A system for
  large-scale machine learning,'' in \emph{USENIX OSDI}, 2016, pp. 265--283.

\bibitem{corrado2014training}
G.~S. Corrado, K.~Chen, J.~A. Dean, S.~Bengio, R.~Monga, and M.~Devin,
  ``Training a model using parameter server shards,'' 7 2014, uS Patent
  8,768,870.

\bibitem{yan2015performance}
F.~Yan, O.~Ruwase, Y.~He, and T.~Chilimbi, ``Performance modeling and
  scalability optimization of distributed deep learning systems,'' in
  \emph{Proceedings of the 21th ACM SIGKDD}.\hskip 1em plus 0.5em minus
  0.4em\relax ACM, 2015, pp. 1355--1364.

\bibitem{shi2018adag}
S.~Shi, Q.~Wang, X.~Chu, and B.~Li, ``A {DAG} model of synchronous stochastic
  gradient descent in distributed deep learning,'' in \emph{ICPADS}, 2018.

\bibitem{awan2017s}
A.~A. Awan, K.~Hamidouche, J.~M. Hashmi, and D.~K. Panda, ``{S-Caffe}:
  Co-designing {MPI} runtimes and {Caffe} for scalable deep learning on modern
  {GPU} clusters,'' in \emph{ACM PPoPP}.\hskip 1em plus 0.5em minus 0.4em\relax
  ACM, 2017, pp. 193--205.

\bibitem{goyal2017accurate}
P.~Goyal, P.~Doll{\'a}r, R.~Girshick, P.~Noordhuis, L.~Wesolowski, A.~Kyrola,
  A.~Tulloch, Y.~Jia, and K.~He, ``Accurate, large minibatch {SGD}: training
  {ImageNet} in 1 hour,'' \emph{arXiv preprint arXiv:1706.02677}, 2017.

\bibitem{jia2018highly}
X.~Jia, S.~Song, S.~Shi, W.~He, Y.~Wang, H.~Rong, F.~Zhou, L.~Xie, Z.~Guo,
  Y.~Yang, L.~Yu, T.~Chen, G.~Hu, and X.~Chu, ``Highly scalable deep learning
  training system with mixed-precision: Training {ImageNet} in four minutes,''
  \emph{NeurIPSW}, 2018.

\bibitem{pjevsivac2007performance}
J.~Pje{\v{s}}ivac-Grbovi{\'c}, T.~Angskun, G.~Bosilca, G.~E. Fagg, E.~Gabriel,
  and J.~J. Dongarra, ``Performance analysis of {MPI} collective operations,''
  \emph{Cluster Computing}, vol.~10, no.~2, pp. 127--143, 2007.

\bibitem{alistarh2017qsgd}
D.~Alistarh, D.~Grubic, J.~Li, R.~Tomioka, and M.~Vojnovic, ``{QSGD}:
  Communication-efficient {SGD} via gradient quantization and encoding,'' in
  \emph{NeurlPS}, 2017, pp. 1707--1718.

\bibitem{wen2017terngrad}
W.~Wen, C.~Xu, F.~Yan, C.~Wu, Y.~Wang, Y.~Chen, and H.~Li, ``Terngrad: Ternary
  gradients to reduce communication in distributed deep learning,'' in
  \emph{NeurlPS}, 2017, pp. 1509--1519.

\bibitem{chen2017adacomp}
C.-Y. Chen, J.~Choi, D.~Brand, A.~Agrawal, W.~Zhang, and K.~Gopalakrishnan,
  ``{AdaComp}: Adaptive residual gradient compression for data-parallel
  distributed training,'' in \emph{AAAI}, 2018, pp. 2827--2835.

\bibitem{lin2017deep}
Y.~Lin, S.~Han, H.~Mao, Y.~Wang, and W.~J. Dally, ``Deep gradient compression:
  Reducing the communication bandwidth for distributed training,'' in
  \emph{ICLR}, 2018.

\bibitem{shi2019distributed}
S.~Shi, Q.~Wang, K.~Zhao, Z.~Tang, Y.~Wang, X.~Huang, and X.~Chu, ``A
  distributed synchronous {SGD} algorithm with global top-k sparsification for
  low bandwidth networks,'' in \emph{ICDCS}, 2019, pp. 2238--2247.

\bibitem{shi2020communication}
S.~Shi, Q.~Wang, X.~Chu, B.~Li, Y.~Qin, R.~Liu, and X.~Zhao,
  ``Communication-efficient distributed deep learning with merged gradient
  sparsification on gpus,'' in \emph{INFOCOM}, 2020.

\bibitem{shi2019aconvergence}
S.~Shi, K.~Zhao, Q.~Wang, Z.~Tang, and X.~Chu, ``A convergence analysis of
  distributed {SGD} with communication-efficient gradient sparsification,'' in
  \emph{IJCAI}, 2019, pp. 3411--3417.

\bibitem{karimireddy2019error}
S.~P. Karimireddy, Q.~Rebjock, S.~Stich, and M.~Jaggi, ``Error feedback fixes
  {SignSGD} and other gradient compression schemes,'' in \emph{ICML}, 2019, pp.
  3252--3261.

\bibitem{lian2017can}
X.~Lian, C.~Zhang, H.~Zhang, C.-J. Hsieh, W.~Zhang, and J.~Liu, ``Can
  decentralized algorithms outperform centralized algorithms? a case study for
  decentralized parallel stochastic gradient descent,'' in \emph{NeurlPS},
  2017, pp. 5330--5340.

\bibitem{DCD-PSGD}
H.~Tang, S.~Gan, C.~Zhang, T.~Zhang, and J.~Liu, ``Communication compression
  for decentralized training,'' in \emph{NeurlPS}, 2018, pp. 7663--7673.

\bibitem{he2016deep}
K.~He, X.~Zhang, S.~Ren, and J.~Sun, ``Deep residual learning for image
  recognition,'' in \emph{CVPR}, 2016, pp. 770--778.

\bibitem{shi2019understanding}
S.~Shi, X.~Chu, K.~C. Cheung, and S.~See, ``Understanding top-k sparsification
  in distributed deep learning,'' \emph{arXiv preprint arXiv:1911.08772}, 2019.

\bibitem{gossipalgorithms}
S.~{Boyd}, A.~{Ghosh}, B.~{Prabhakar}, and D.~{Shah}, ``Gossip algorithms:
  design, analysis and applications,'' in \emph{INFOCOM}, 2005, pp. 1653--1664.

\bibitem{RandomizedGossipAlgorithms}
S.~Boyd, A.~Ghosh, B.~Prabhakar, and D.~Shah, ``Randomized gossip algorithms,''
  \emph{IEEE/ACM Trans. Netw.}, vol.~14, pp. 2508--2530, 2006.

\bibitem{karp1972reducibility}
R.~M. Karp, ``Reducibility among combinatorial problems,'' in \emph{Complexity
  of computer computations}.\hskip 1em plus 0.5em minus 0.4em\relax Springer,
  1972, pp. 85--103.

\bibitem{randomrumorspreading}
R.~{Karp}, C.~{Schindelhauer}, S.~{Shenker}, and B.~{Vocking}, ``Randomized
  rumor spreading,'' in \emph{FOCS}, 2000, pp. 565--574.

\bibitem{edmonds_1965}
J.~Edmonds, ``Paths, trees, and flowers,'' \emph{Canadian Journal of
  Mathematics}, vol.~17, p. 449–467, 1965.

\bibitem{renggli2019sparcml}
C.~Renggli, S.~Ashkboos, M.~Aghagolzadeh, D.~Alistarh, and T.~Hoefler,
  ``Sparcml: High-performance sparse communication for machine learning,'' in
  \emph{SC}, 2019, pp. 1--15.

\bibitem{mcmahan2016communication}
B.~McMahan, E.~Moore, D.~Ramage, S.~Hampson, and B.~A. y~Arcas,
  ``Communication-efficient learning of deep networks from decentralized
  data,'' in \emph{AISTAT}, 2017, pp. 1273--1282.

\bibitem{dekel2012optimal}
O.~Dekel, R.~Gilad-Bachrach, O.~Shamir, and L.~Xiao, ``Optimal distributed
  online prediction using mini-batches,'' \emph{JMLR}, vol.~13, no. Jan, pp.
  165--202, 2012.

\bibitem{lecun2010mnist}
Y.~LeCun, C.~Cortes, and C.~Burges, ``{MNIST} handwritten digit database,''
  \emph{AT\&T Labs [Online]. Available: http://yann. lecun. com/exdb/mnist},
  vol.~2, p.~18, 2010.

\bibitem{krizhevsky2010cifar}
A.~Krizhevsky, V.~Nair, and G.~Hinton, ``The {CIFAR-10} dataset,'' vol.~55,
  2014.

\bibitem{seide20141}
F.~Seide, H.~Fu, J.~Droppo, G.~Li, and D.~Yu, ``1-bit stochastic gradient
  descent and its application to data-parallel distributed training of speech
  {DNNs},'' in \emph{INTERSPEECH}, 2014.

\bibitem{hubara2017quantized}
I.~Hubara, M.~Courbariaux, D.~Soudry, R.~El-Yaniv, and Y.~Bengio, ``Quantized
  neural networks: Training neural networks with low precision weights and
  activations,'' \emph{JMLR}, vol.~18, no.~1, pp. 6869--6898, 2017.

\bibitem{wangni2018gradient}
J.~Wangni, J.~Wang, J.~Liu, and T.~Zhang, ``Gradient sparsification for
  communication-efficient distributed optimization,'' in \emph{NeurlPS}, 2018,
  pp. 1299--1309.

\bibitem{zhao2018federated}
Y.~Zhao, M.~Li, L.~Lai, N.~Suda, D.~Civin, and V.~Chandra, ``Federated learning
  with non-iid data,'' \emph{arXiv preprint arXiv:1806.00582}, 2018.

\bibitem{shi2020layer}
S.~Shi, Z.~Tang, Q.~Wang, K.~Zhao, and X.~Chu, ``Layer-wise adaptive gradient
  sparsification for distributed deep learning with convergence guarantees,''
  in \emph{ECAI}, 2020.

\bibitem{sirb2018decentralized}
B.~Sirb and X.~Ye, ``Decentralized consensus algorithm with delayed and
  stochastic gradients,'' \emph{SIAM Journal on Optimization}, vol.~28, no.~2,
  pp. 1232--1254, 2018.

\end{thebibliography}
\end{document}